\documentclass[journal]{IEEEtran}

\usepackage{cite}

\usepackage[fontsize=0.02\paperwidth,angle=0,vpos=0.195\paperheight,firstpageonly=true]{draftwatermark}

\usepackage{lipsum}

\usepackage{amsmath}
\usepackage{amsfonts}
\usepackage{amssymb}
\usepackage{mathtools}

\usepackage{multirow}
\usepackage{booktabs}

\usepackage{xspace}

\usepackage{subcaption}
\usepackage{svg}

\usepackage{stfloats}

\usepackage{xcolor}

\usepackage{algorithm}
\usepackage{algorithmic}
\usepackage{nicefrac}

\usepackage{tikz}

\usepackage{hyperref}
\usepackage{url}

\usepackage{soul}

\newsavebox{\bigimage}

\newcommand{\change}[1]{#1}

\DeclareMathOperator*{\argmin}{arg\,min}

\newcommand{\cvec}[1]{\boldsymbol{#1}}
\newcommand{\svec}[1]{\mathbf{#1}}

\newcommand{\dif}{\mathop{}\!\mathrm{d}}

\newcommand{\ddp}{\textsc{DDP}\xspace}
\newcommand{\ppo}{\textsc{PPO}\xspace}
\newcommand{\currot}{\mbox{\textsc{CurrOT}}\xspace}
\newcommand{\lowdimcurrot}{\mbox{$\mathrm{\textsc{CurrOT}}_{L}$}\xspace}
\newcommand{\affinecurrot}{\mbox{$\mathrm{\textsc{CurrOT}}_{A}$}\xspace}
\newcommand{\newcurrot}{\mbox{$\mathrm{\textsc{CurrOT}}_{AO}$}\xspace}

\definecolor{c3red}{RGB}{197,58,50}
\definecolor{c2green}{RGB}{81,158,62}
\definecolor{c1orange}{RGB}{239,134,54}
\definecolor{c0blue}{RGB}{58,115,172}

\begin{document}

\title{Tracking Control for a Spherical Pendulum via Curriculum Reinforcement Learning
}

\author{Pascal Klink$^{1}$, Florian Wolf$^{1}$, Kai Ploeger$^{1}$, Jan Peters$^{1,3}$, and Joni Pajarinen$^{2}$
\thanks{$^{1}$P.~Klink, F.~Wolf, K.~Ploeger and J.~Peters are with the Intelligent Autonomous Systems Group at the Technical University of Darmstadt, Germany. Correspondence to:  {\tt\small pascal.klink@tu-darmstadt.de}}%
\thanks{$^{2}$J.~Pajarinen is with the Department of Electrical Engineering and Automation at Aalto University, Finland.}
\thanks{$^{3}$J.~Peters is also with the German Research Center for AI (Research Department: Systems AI for Robot Learning), Hessian.AI and the Centre of Cognitive Science.}}%

\markboth{SUBMITTED TO IEEE TRANSACTIONS ON ROBOTICS}%
{Klink \MakeLowercase{\textit{et al.}}: Tracking Control for a Spherical Pendulum via Curriculum Reinforcement Learning}
\SetWatermarkColor[rgb]{{1, 0.6, 0.6}}
\SetWatermarkText{This work has been submitted to the IEEE for possible publication. \vspace{-3pt} \\
Copyright may be transferred without notice, after which this version may no longer be accessible.}

\maketitle

\begin{abstract}
Reinforcement Learning (RL) allows learning non-trivial robot control laws purely from data. However, many successful applications of RL have relied on ad-hoc regularizations, such as hand-crafted curricula, to regularize the learning performance. In this paper, we pair a recent algorithm for automatically building curricula with RL on massively parallelized simulations to learn a tracking controller for a spherical pendulum on a robotic arm via RL. Through an improved optimization scheme that better respects the non-Euclidean task structure, we allow the method to reliably generate curricula of trajectories to be tracked, resulting in faster and more robust learning compared to an RL baseline that does not exploit this form of structured learning. The learned policy matches the performance of an optimal control baseline on the real system, demonstrating the potential of curriculum RL to jointly learn state estimation and control for non-linear tracking tasks.
\end{abstract}


\section{Introduction}

\begin{figure}[t!]
    \centering
    \includegraphics[width=0.23\textwidth]{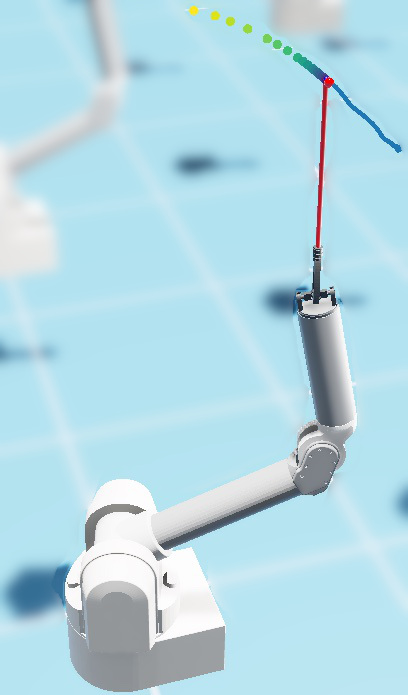} \includegraphics[width=0.23\textwidth]{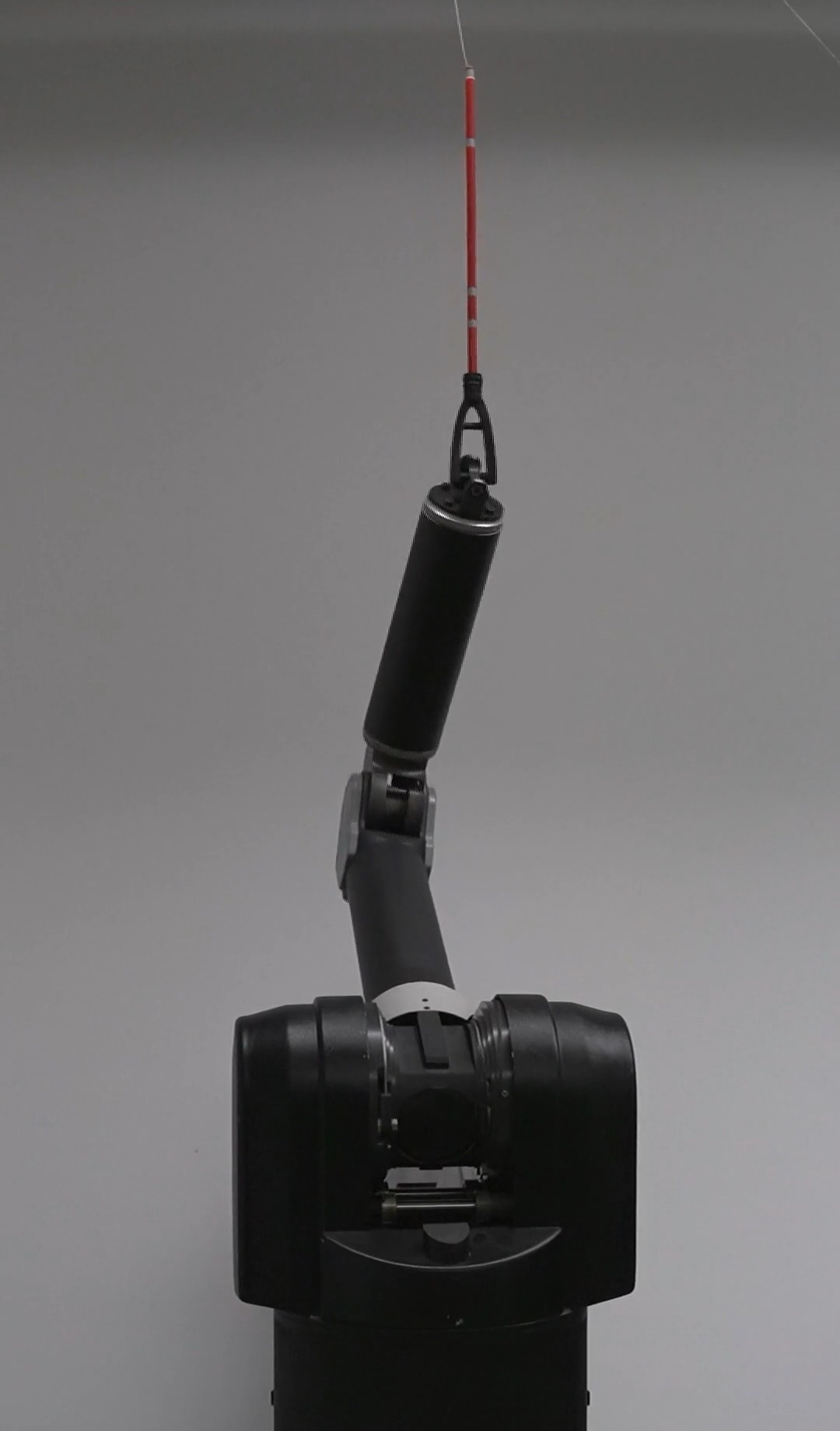}
    \caption{An image of our simulation (left) and robot environment (right) of the spherical pendulum tracking task. The eight-shaped target trajectories evolving in three spatial dimensions require coordinated control to simultaneously balance the pendulum and achieve good tracking performance. The pendulum is mounted to a Barrett WAM robotic arm and is tracked by an Optitrack system. In the left image, the colored dots visualize the upcoming target trajectory to be followed, and the blue line visualizes the achieved trajectory.}
    \label{fig:spherical:pull-figure}
\end{figure}

\noindent Due to a steady increase in available computation over the last decades, reinforcement learning (RL) \cite{sutton1998introduction} has been applied to increasingly challenging learning tasks both in simulated \cite{mnih2015human,silver2016mastering} and robotic domains \cite{levine2018learning,akkaya2019solving,rudin2022learning}. Learning control of non-trivial systems via reinforcement learning (RL) is particularly appealing when dealing with partially observable systems and high-dimensional observations such as images, or if quick generalization to multiple related tasks is desired. \\
In this paper, we provide another demonstration of the potential of reinforcement learning to find solutions to a non-trivial control task that has, to the best of our knowledge, not been tackled using learning-based methods. More precisely, we focus on the tracking control of a spherical pendulum attached to a four degrees-of-freedom Barrett Whole Arm Manipulator (WAM) \cite{barrett2023wam}, as shown in Figure \ref{fig:spherical:pull-figure}. The partial observability of the system arising from access to only positional information paired with an inherently unstable, underactuated system and non-trivial kinematics results in a challenge for modern reinforcement learning algorithms. \\
With reinforcement learning being applied to increasingly demanding learning tasks such as the one presented in this paper, different strategies for improving learning performance, such as guiding the learning agent through highly shaped and -informative reward functions \cite{ng1999policy,gupta2022unpacking}, have evolved. In this paper, we improve the training performance of the learning agent via curricula, i.e., tailored sequences of learning tasks that adapt the environment's complexity to the capability of the learning agent. 
For the considered tracking task, we adapt the complexity via the target trajectories that are to be tracked by the controller, starting from small deviations from an initial position and progressing to a set of eight-shaped target trajectories requiring the robot to move in all three dimensions. \\
Scheduling the complexity of the learning tasks is subject to ongoing research \cite{narvekar2020curriculum}, and solutions to this problem are motivated from different perspectives, such as two-player games \cite{sukhbaatar2018intrinsic} or the maximization of intrinsic motivation \cite{baranes2010intrinsically}. \\
In this paper, we generate the curriculum of tasks using the \currot algorithm \cite{klink2022curriculum}, which defines the curriculum as a constrained interpolation between an initial- and desired distribution of training tasks and is well-suited to our goal of directing learning to a set of target trajectories. Applications of \currot have so far relied on training tasks that can be represented in a low-dimensional vector space. We create a curriculum over desired trajectories, a high-dimensional space of learning tasks, allowing us to benchmark the \currot algorithm in this unexplored setting. \\
We demonstrate that the sampling-based optimization scheme of \currot that drives the evolution of the learning tasks faces challenges in high-dimensional scenarios. Furthermore, the default assumption of a Euclidean distance on the vector space of learning tasks can lead to curricula that do not facilitate learning. Addressing both pitfalls, we obtain robust convergence to the target distribution of tasks, resulting in a tracking controller that can be applied to the real system.  
\textbf{Contributions:}
\begin{itemize}
    \item We demonstrate a simulation-based approach for learning tracking controllers for an underactuated, partially observable, and highly unstable non-linear system that directly transfer to reality.
    \item Our approach includes a curriculum reinforcement learning method that reliably works with high-dimensional task spaces equipped with Mahalanobis distances, such as trajectories, commonly encountered in robotics.
    \item Through ablations, we confirm the robustness of our method and provide insights into the importance of the policy structure for generalization in tracking tasks.
\end{itemize}

\section{Related Work}

\noindent As of today, there exist many demonstrations of applying reinforcement learning (RL) to real-world robotic problems, ranging from locomotion \cite{rudin2022learning,chen2022learning,li2023robust} to object manipulation \cite{gu2017deep,levine2018learning,liu2023safe}, where the RL agents need to process high-dimensional observations, such as images \cite{levine2018learning} or grids of surface height measurements \cite{rudin2022learning} in order to produce appropriate actions. The RL agent typically controls the robot via desired joint positions \cite{rudin2022learning,li2023robust}, joint position deltas \cite{levine2018learning}, joint velocities \cite{gu2017deep}, or even joint torques \cite{li2023robust}. Depending on the application scenario, actions are restricted to a manifold of save actions \cite{gu2017deep,liu2023safe}. \\
\noindent \textbf{Spherical Pendulum:} Inverted pendulum systems have been investigated since the 1960s \cite{lundberg2010history} as an archetype of an inherently unstable system and are a long-standing evaluation task for reinforcement learning algorithms \cite{selfridge1985training}, with swing-up and stabilization tasks successfully solved on real systems via RL \cite{kober2008policy,lutter2022continuous}. Other learning-based approaches tune linear quadratic regulators (LQRs) and PID controllers in a data-driven manner to successfully stabilize an inverted pendulum mounted on a robotic arm \cite{marco2016automatic,doerr2017model}. The extension of the one-dimensional inverted pendulum task to two dimensions has been widely studied in the control community, resulting in multiple real-world applications in which the pendulum has been mounted either to an omnidirectional moving base \cite{kao2013balancing,kao2017tracking}, a platform driven via leading screws \cite{yang2000stabilization}, a SCARA robotic arm \cite{sprenger1998balancing}, or a seven degrees-of-freedom collaborative robotic arm \cite{vu2021fast}. The controllers for these systems were synthesized either via linear controller design in task space \cite{sprenger1998balancing}, a time-variant LQR around pre-planned trajectories \cite{vu2021fast}, linear output regulation \cite{kao2017tracking}, sliding-mode control \cite{kao2013balancing}, or feedback linearization \cite{yang2000stabilization}. In these approaches, the control laws assumed observability of the complete state, requiring specially designed pendulum systems featuring joint encoders or magneto-resistive sensors and additional processing logic to infer velocities. \\
In this paper, we learn tracking control of a spherical pendulum on a robotic arm from position-only observations via reinforcement learning. To the best of our knowledge, this has not yet been achieved, and we believe that the combination of non-trivial kinematics, underactuation, and partial observability is a good opportunity to demonstrate the capabilities of modern deep RL agents. \\
\noindent \textbf{Curriculum Reinforcement Learning:} The complexity of this learning task provides an opportunity to utilize methods from the field of curriculum reinforcement learning \cite{narvekar2020curriculum}. These methods improve the learning performance of RL agents in various application scenarios \cite{silver2016mastering,akkaya2019solving,rudin2022learning} by adaptively modifying environment aspects of a contextual- \cite{hallak2015contextual} or, more generally, a configurable Markov Decision Process \cite{metelli18configurable}. As do their application scenarios, motivations for and realizations of these algorithms differ widely, e.g., in the form of two-player games \cite{sukhbaatar2018intrinsic,dennis2020emergent}, approaches that maximize intrinsic motivation \cite{baranes2010intrinsically,portelas2019teacher}, or as interpolations between task distributions \cite{chen2021variational,klink2022curriculum}. We will focus on the \currot algorithm \cite{klink2022curriculum} belonging to the last category of approaches, as it is well suited for our goal of tracking a specific set of target trajectories and has so far been applied to rather low-dimensional settings, allowing us to extend its application scenarios to the high-dimensional space of trajectories faced here.

\section{Reinforcement Learning System}

\noindent In this section, we describe the trajectory tracking task, its simulation in IsaacSim \cite{nvidia2022isaacsim}, and the curriculum learning approach \cite{klink2022curriculum} we utilized to speed up learning in this environment.

\subsection{Simulation Environment and Policy Representation}

\begin{figure}[t]
    \centering
    \includegraphics{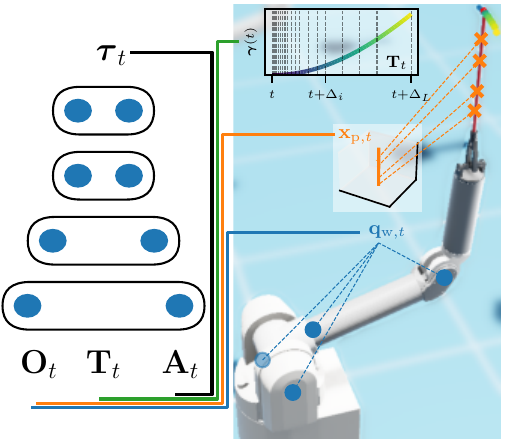}
    \caption{The policy is a standard feedforward neural network with $[1024, 512, 256, 256]$ hidden layers that observes a history $\svec{O}_t$ of joint positions $\svec{q}_{\text{w},t}$ and pole directions $\svec{x}_{p,t}$, a history $\svec{A}_t$ of past actions $\cvec{\tau}_t$, and a lookahead $\svec{T}_t$ of the trajectory $\cvec{\gamma}(t)$ to be followed.}
    \label{fig:spherical:policy_architecture}
\end{figure}

\noindent As shown in Figure \ref{fig:spherical:pull-figure}, we aim to learn a tracking task of a spherical pendulum that is mounted on a four degrees-of-freedom Barrett Whole Arm Manipulator (WAM) \cite{barrett2023wam} via a 3D printed universal joint\footnote{We designed the universal joint such that it has a large range of motion. Furthermore, the use of skateboard bearings resulted in low joint friction.}. The robot can be approximately modeled as an underactuated rigid body system
\begin{align}
    \svec{M}(\svec{q}) \ddot{\svec{q}} = \svec{c}(\svec{q}, \dot{\svec{q}}) + \svec{g}(\svec{q}) + \cvec{\tau}_{\textrm{pad}} \label{eq:spherical:dynamics}
\end{align}
with six degrees of freedom $\svec{q} = [\svec{q}_{\textrm{w}}\  \svec{q}_{\textrm{p}}] \in \mathbb{R}^6$ that represent the joint positions of the Barrett WAM ($\svec{q}_{\textrm{w}}$) and the pendulum ($\svec{q}_{\textrm{P}}$), and four control signals $\cvec{\tau} \in \mathbb{R}^4$ that drive the joints of the Barrett WAM, where $\cvec{\tau}_{\textrm{pad}} = [\cvec{\tau}\ 0\ 0]$ appends the (always zero) controls for the non-actuated universal joint of the spherical pendulum. The universal joint does not possess any encoders, and we can infer the state of the pole only through position measurements provided by an OptiTrack system \cite{natural2023optitrack} at $120$ Hz. Hence, albeit the Barrett WAM can be controlled at $500$ Hz and delivers updates on its joint positions at the same frequency, we run the control law only at $125$ Hz due to the OptiTrack frequency. In the following, we denote a variable's value at a discrete time index as $x_t$ and the value at arbitrary continuous time as $x(t)$. We learn a tracking control law for following desired trajectories $\cvec{\gamma}{:} [t_s, t_e] {\mapsto} \mathbb{R}^3$ of the pendulum tip from a fixed initial configuration $\svec{q}_{\textrm{w},0}$. The control law generates torques on top of a gravity compensation term $\svec{g}(\svec{q}_{\textrm{w}})$ based on a history of positional observations, applied torques, and information about the desired trajectory $\cvec{\gamma}$
\begin{align}
    \cvec{\tau}_t &{=} \pi(\svec{O}_t, \svec{A}_t, \svec{T}_t) {+} \svec{g}(\svec{q}_{\text{w},\change{t}}) & \svec{O}_t &{=} \{\svec{o}_{t - i} | i \in [0, K{-}1]\} \nonumber \\
    \svec{T}_t &{=} \{ (\cvec{\gamma}_{t {+} \Delta_i}, \dot{\cvec{\gamma}}_{t {+} \Delta_i}) | i {\in} [1,L] \} & \svec{A}_t &{=} \{\cvec{\tau}_{t-i} | i {\in} [1,K] \},
\end{align}
where $K{=}15$, $L{=}20$, and the $\Delta_i$'s are spread out over the interval $[0,1.04]$ (Figure \ref{fig:spherical:policy_architecture}) to capture both the immediately upcoming positions and velocities of $\cvec{\gamma}(t)$ as well as the future behavior of the trajectory. An observation $\svec{o}_t$ is given by the joint position of the Barrett WAM $\svec{q}_{\text{w},t}$ as well as a three-dimensional unit-vector $\svec{x}_{\text{p}, t} {\in} \mathbb{R}^3$ that represents the orientation of the pole (Figure \ref{fig:spherical:policy_architecture}). In simulation, we compute this vector using the difference between the pendulum tip $\svec{x}_{\text{tip}, t}$ and the pendulum base $\svec{x}_{\text{base}, t}$. On the real system, we compute this vector from Optitrack measurements of four points on the pendulum. We reconstruct neither the pendulum joint positions $\svec{q}_{p,t}$ nor the joint velocities $\dot{\svec{q}_{\change{t}}}$ since this information is implicitly contained in the observation- and action histories $\svec{O}_t$ and $\svec{A}_t$.  \\
We learn $\pi$ via the \textit{proximal policy optimization} (\ppo) algorithm implemented in the \textit{RL Games} library \cite{rl-games2021}. This choice is motivated by our use of the IsaacSim simulation environment \cite{nvidia2022isaacsim}, which allows us to simulate a large number of environments in parallel on a single GPU\footnote{We used $2048$ parallel environments for learning.}. The chosen \ppo implementation is designed to leverage this parallel simulation during training. Screenshots of the simulation environment are shown in Figures \ref{fig:spherical:pull-figure} and \ref{fig:spherical:policy_architecture}. The trajectories evolve over a total duration of $12$ seconds, resulting in $12 {\cdot} 125 {=} 1500$ steps per episode. The reward function at a given time-step $t$ mainly penalizes tracking failures and additionally regularizes excessive movement of the robot
\begin{align}
    r(\svec{q}_t, \cvec{\tau}_t) {=} \begin{cases} 
    -\frac{\alpha}{1 - \gamma}, \qquad \qquad \qquad \qquad \qquad \text{ if } \text{tipped}(\svec{q}_t) \\
    1 - 1000 \| \cvec{\gamma}_t {-} \svec{x}_{\textrm{tip},t} \|_2^2 {-} 1e^{-1} \|\dot{\svec{q}}_{\textrm{w}\change{,t}}\|_2^2 \\
    \quad {-} 1e^{-1} \| \svec{q}_{\textrm{w},t} {-} \svec{q}_{\textrm{w},0} \|_2^2 - 1e^{-3} \| \cvec{\tau}_t \|_2^2, \text{ else.}
    \end{cases} \label{eq:spherical:reward}
\end{align}
The function $\text{tipped}(\svec{q}_t)$ returns true if either $|\svec{q}_{\text{p},t}| {\geq} 0.5 \pi$ or if the $z$-coordinate of the pendulum tip $\svec{x}_{\text{tip},t}$ is less than five centimeters above the z-coordinate of the pendulum base  $\svec{x}_{\text{base},t}$. The episode ends if $\text{tipped}(\svec{q}_t)$ evaluates to true. The large amplification of the tracking error is required since $\| \cvec{\gamma}_t {-} \svec{x}_{\textrm{tip},t} \|_2$ is measured in meters. With the chosen amplification, a tracking error of three centimeters leads to a penalty of $-0.9$. In our experiments, we use a discount factor of $\gamma {=} 0.992$ \change{and evaluate the learning system for multiple $\alpha$}.

\subsection{Facilitating Sim2Real Transfer}

\noindent To enable successful transfer from simulation to reality, we first created a rigid-body model of the Barrett WAM (Eq. \ref{eq:spherical:dynamics}) based on the kinematic and inertial data sheets from Barrett Technology \cite{barrett2023wam} in the MuJoCo physics simulator \cite{todorov2012mujoco}. We chose the MuJoCo simulator for initial investigations since it allows us to more accurately model the actuation of the Barrett WAM via tendons and differentials\footnote{IsaacSim also has support for tendon modelling. However, this support is significantly more restricted at the moment, preventing to recreate the tendon structure of the Barrett WAM in simulation.}. Opposed to the simplified model (\ref{eq:spherical:dynamics}), this more faithful model of the Barrett WAM requires an extended state space $\svec{q}_{\text{ext}} {=} [\svec{q}_{\text{w}} \ \svec{q}_{\text{r}} \ \svec{q}_{\text{p}}] {\in} \mathbb{R}^{10}$, in which the joint- $\svec{q}_{\text{w}}$ and rotor positions $\svec{q}_{\text{r}}$ of the Barrett WAM are coupled via tendons that transfer the torques generated at the rotors to the joints (and vice versa). The joint encoders of the WAM are located at the rotors, and hence, we can only observe $\svec{q}_{\text{r}}$, which may differ from $\svec{q}_{\text{w}}$ depending on the stiffness of the tendons. During our initial evaluations, we found that modeling this discrepancy between measured- and real joint positions as well as delayed actions (approximated as an exponential filter)
\begin{align}
    \tilde{\cvec{\tau}}_{t} = \change{\cvec{\omega}} \odot \tilde{\cvec{\tau}}_{t-1} + (\svec{1} - \change{\cvec{\omega}}) \odot \cvec{\tau}_t, \quad \change{\cvec{\omega} \in [0,1]^4,}
\end{align}
where $\odot$ represents the element-wise multiplication of vectors and $\svec{1}$ is a vector of all ones, were required to achieve stable behavior of the learned policy on the real system. When not modeling these effects, the actions generated by the learned policies resulted in unstable feedback loops. A final extension to the model is given by simulating a Stribeck-like behavior of friction by compensating the coulomb friction modeled by MuJoCo
\begin{align}
    \tilde{\cvec{\tau}}_{a,t} = \tilde{\cvec{\tau}}_{\change{t}} + \svec{c} \odot \tanh(\cvec{\beta} \odot \dot{\svec{q}}_{\text{w}}),
\end{align}
where $\svec{c} \change{{\in} \mathbb{R}^4_{\geq0}}$ is the coefficient of coulomb friction simulated by MuJoCo, $\cvec{\beta} \change{{\in} \mathbb{R}^4_{\geq0}}$ is the reduction of friction due to movement. Having completed our model, we then adjusted the tendon stiffness, rotor armature, damping, coulomb friction $\svec{c}$, as well as $\change{\cvec{\omega}}$ and $\cvec{\beta}$ using trajectories from the real system. \\
\change{Given the lack of possibilities to model the tendon drives of the Barret WAM in IsaacSim, we simulate the robot without tendons and model the discrepancies between $\svec{q}_{\text{r}}$ observed by the policy and $\svec{q}_{\text{w}}$ by a simple spring-damper model}
\begin{align}
    \ddot{\svec{q}}_{\text{r}} = \svec{K}_P (\svec{q}_{\text{\change{w}}} - \svec{T}_{\svec{q}} \svec{q}_{\text{r}}) + \svec{K}_D (\dot{\svec{q}}_{\text{\change{w}}} - \svec{T}_{\svec{q}} \dot{\svec{q}}_{\text{r}}) + \svec{T}_{\cvec{\tau}} \change{\tilde{\cvec{\tau}}}
\end{align}
where $\svec{T}_\svec{q}, \svec{T}_{\cvec{\tau}} {\in} \mathbb{R}^{4\times4}$ model the transformation of joint position and -torques via the tendons and $\svec{K}_P, \svec{K}_D {\in} \mathbb{R}^{4\times4}$ model the spring-damper properties of the tendons. \\
Given the policy's reliance on Optitrack measurements of the pendulum, which are exchanged over the network, we measured the time delays arising from the communication over the network stack. We then modeled these delays in the simulation, as detailed in Appendix \ref{app:spherical:network}. \\
During learning, we randomize the masses within $75\%$ and $125\%$ of their nominal values and randomize damping and coulomb friction within $50\%$ and $150\%$ of their nominal values. Additionally, we add zero-mean Gaussian distributed noise with a standard deviation of $0.005$ to the actions generated by the agent, which are normalized between $-1$ and $1$. The observations are corrupted by uniform noise with\change{in} $[-0.01, 0.01]$. Finally, the amount of action delay is also randomized by sampling $\change{\cvec{\omega}}$ from $[0.5, 0.9]$, and $\cvec{\beta}$ is set to zero $25\%$ of the time and sampled from $[0,100]$ otherwise.

\subsection{Trajectory Representations}

\noindent We represent the target trajectories $\cvec{\gamma}{:} [t_s,t_e] {\mapsto} \mathbb{R}^3$ via a constrained three-dimensional LTI system that is driven by a sequence of jerks (time-derivatives of accelerations)
\begin{align}
    &\forall t \in [t_s, t_e]: \cvec{\gamma}(t) \in \mathcal{P} \label{eq:spherical:pos-constraint} \\
    &\forall t \in [t_s, t_e]: \left\| \frac{\dif^3}{\dif t^3}\cvec{\gamma}(t) \right\|_2 \leq j_{\text{UB}} \label{eq:spherical:jerk-constraint} \\
    &\cvec{\gamma}(t_s) {=} \cvec{\gamma}(t_e) \quad \dot{\cvec{\gamma}}(t_s) {=} \dot{\cvec{\gamma}}(t_e) {=} 0 \quad \ddot{\cvec{\gamma}}(t_s) {=} \ddot{\cvec{\gamma}}(t_e) {=} 0 \label{eq:spherical:closing-constraint}
\end{align}
with a convex set $\mathcal{P} {\subset} \mathbb{R}^3$ of allowed positions. We model the LTI system as three individual triple integrator models. For simplicity of exposition, we focus on only one of the three systems, i.e., $\gamma{:} [t_s,t_e] {\mapsto} \mathbb{R}$. The full system is obtained by simple ``concatenation'' of three copies of the following system
\begin{align}
    \dot{\svec{x}}(t) &= \svec{A} \svec{x}(t) + \svec{B} u(t) \quad \svec{A} {=} \begin{bmatrix} 0 & 1 & 0 \\
    0 & 0 & 1 \\
    0 & 0 & 0 \end{bmatrix} \quad \svec{B} {=} \begin{bmatrix} 0 \\ 0 \\ 1 \end{bmatrix} \label{eq:spherical:lti-sys}
\end{align}
with $x_i(t) = \frac{\dif^{i-1}}{\dif t^{i-1}}\gamma(t)$ and $u(t) = \frac{\dif^3}{\dif t^3}\gamma(t)$. To represent the trajectories as some finite-dimensional vectors $\svec{u} \in \mathbb{R}^K$, we assume that the control trajectory of jerks $u(t)$ is piece-wise constant
\begin{align*}
    u(t) = \sum_{k=1}^K u_k 1_k(t),\quad 1_k(t) {=} \begin{cases} 
    1,\ \textrm{if }  t_{k-1} \leq t < t_k \\
    0,\ \textrm{else}
    \end{cases}
\end{align*}
with $t_0{=}t_s$ and $t_K{=}t_e$. This assumption allows us to represent $\svec{x}(t)$ at time $t$ as a linear combination \change{of the initial system state and the piece-wise constant jerks}
\begin{align}
    \svec{x}(t) {=} &\cvec{\Phi}(t_s, t) \svec{x}(t_s)\  + \nonumber \\[-1.7em]
    & \change{\hspace{-0.33em} \underbrace{\begin{bmatrix} \cvec{\psi}(t_s, t_1, t)\, \cvec{\psi}(t_1, t_2, t) \ldots \cvec{\psi}(t_{K{-}1}, t_K, t) \end{bmatrix}}_{\cvec{\Psi}(t) \in \mathbb{R}^{3 \times K}} \hspace{-3pt} \underbrace{\begin{bmatrix} u_{1} \\ u_{2} \\ \vdots \\ u_{K} \end{bmatrix}}_{\svec{u} \in \mathbb{R}^K} \hspace{-3pt} .} \label{eq:spherical:closed-form-lti}
\end{align}
We derive $\cvec{\Phi}$ and $\cvec{\psi}$ in the appendix. With the closed-form solution (\ref{eq:spherical:closed-form-lti}), we can rewrite Constraint (\ref{eq:spherical:closing-constraint}) as a system of three linear equations
\begin{align}
    &\svec{x}(t_e) = \svec{\Phi}(t_s, t_e) \svec{x}\change{(t_s)} {+} \svec{\Psi}\change{(t_e)} \svec{u} \nonumber \\ 
    \Leftrightarrow &\svec{x}(t_e) - \svec{\Phi}(t_s, t_e) \svec{x}\change{(t_s)} {=} \change{\svec{\Psi}\change{(t_e)} \svec{u} \Leftrightarrow}\ \svec{0} {=} \svec{\Psi}\change{(t_e)} \svec{u}. \label{eq:spherical:linear-closing-constraint}
\end{align}
We know that $\svec{x}(t_e) - \cvec{\Phi}(t_s, t_e) \svec{x}\change{(t_s)} = \svec{0}$ due to the form of $\cvec{\Phi}(t_s, t_e)$, and since our initial state $\svec{x}(t_s)$ is, per definition, given by $\svec{x}_s {=} [\gamma(t_s)\, 0\, 0]$. We can hence represent all trajectories that fulfill Constraint (\ref{eq:spherical:closing-constraint}) in a $(K{-}3)$-dimensional basis of the kernel $\text{ker}(\cvec{\Psi}\change{(t_e)})$. We refer to vectors in this kernel as $\tilde{\svec{u}} \in \mathbb{R}^{K-3}$. The two remaining constraints (\ref{eq:spherical:pos-constraint}) and (\ref{eq:spherical:jerk-constraint}) specify a convex set in $\text{ker}(\cvec{\Psi}\change{(t_e)})$. As described in the next section, generating a curriculum over trajectories will require sampling in an $\epsilon$-ball around a given \change{kernel element $\tilde{\svec{u}}$} in the convex set, which we perform using simple rejection sampling.

\subsection{Curriculum Reinforcement Learning}
\label{sec:spherical:currot}

\noindent By now, we can represent target trajectories $\cvec{\gamma}(t)$ via a \change{vector} $\mathbf{c} {=} \left[ \tilde{\mathbf{u}}_1\ \tilde{\mathbf{u}}_2\ \tilde{\mathbf{u}}_3 \right] {\in} \mathbb{R}^{3 (K -3)}$ that represent the trajectory behavior in the three spatial dimensions. We will treat $\cvec{\gamma}(t)$ and $\mathbf{c}$ interchangeably for the remainder of this paper and refer to $\mathbf{c}$ as context or task, following the wording in \cite{klink2022curriculum}. We are interested in learning a policy $\pi$ that performs well on a target distribution $\mu(\cvec{\gamma}) {=} \mu(\svec{c})$ over trajectories. To facilitate learning, we use the curriculum method \currot \cite{klink2022curriculum}, which creates a curriculum of task distributions $p_i(\svec{c})$ by iteratively minimizing their Wasserstein distance $\mathcal{W}_2(p, \mu)$ to the target distribution $\mu(\svec{c})$ \change{under a given distance function $d(\svec{c}_1, \svec{c}_2)$ and} subject to a performance constraint
\begin{align}
    \argmin_{p} \mathcal{W}_2(p, \mu) \label{eq:spherical:currot} \ \  \text{s.t. } p(\mathcal{V}(\pi, \delta)) = 1,
\end{align}
where the set $\mathcal{V}(\pi, \delta) = \left\{ \svec{c} \in \mathcal{C} \middle| J(\pi, \svec{c}) \geq \delta \right\}$ is the set of contexts $\svec{c} \in \mathcal{C}$ in which the agent achieves a performance $J(\pi,\svec{c}) = \mathbb{E}_{\pi} \left[ \sum_{t=0}^{\infty} \gamma^t r(\svec{q}_t,\cvec{\tau}_t) \right]$ of at least $\delta$. We refer to \cite{klink2022curriculum} for the precise definition and derivation of the algorithm and, for brevity, only state the resulting algorithm. \\ 
The task distribution $p_i(\svec{c})$ is represented by a set of $N$ particles, i.e., $\hat{p}_i(\svec{c}) = \frac{1}{N} \sum_{n=1}^N \delta_{\svec{c}_{p_i,n}}(\svec{c})$ with $\delta_{\svec{c}_{\text{ref}}}(\svec{c})$ being the Dirac distribution on $\svec{c}_{\text{ref}}$. Each particle is updated \change{by minimizing the distance $d(\svec{c}, \svec{c}_{\mu, \phi(n)})$ to a target particle~$\svec{c}_{\mu, \phi(n)}$}
\begin{align}
    \min_{\svec{c} \in \mathcal{C}} d(\svec{c}, \svec{c}_{\mu,\change{\phi(n)}}) \ \  \text{s.t. } &\hat{J}(\pi, \svec{c}) \geq \delta \ \  d(\svec{c}, \svec{c}_{p_i,n}) \leq \epsilon, \label{eq:spherical:currot-ind-opt}
\end{align}
where $\hat{J}(\pi, \svec{c})$ is a prediction of $J(\pi, \svec{c})$ using Nadaraya-Watson kernel regression \cite{nadaraya1964estimating}
\begin{align}
    \hat{J}(\pi, \svec{c}) {=} \frac{\sum_{l=1}^L K_h(\svec{c}, \svec{c}_l) \change{J}_l}{\sum_{l=1}^L K_h(\svec{c}, \svec{c}_l)}, \; K_h(\svec{c}, \svec{c}_l) {=} \exp\left(-\frac{d(\svec{c}, \svec{c}_l)^2}{2 h^2}\right).
\end{align} \begin{figure}[t]
    \centering
    \includegraphics{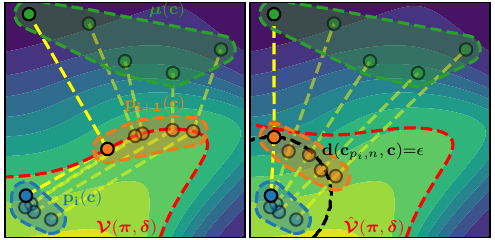}
    \caption{Task sampling scheme used by \currot. (Left) A particle-based representation $\hat{p}_i(\svec{c})$ of the task distribution $p_i(\svec{c})$ is updated to minimize the Wasserstein distance $\mathcal{W}_2(\hat{p}_i, \hat{\mu})$ while keeping all particles in the feasible set $\mathcal{V}(\pi, \delta)$ of tasks in which agent $\pi$ achieves a performance of at least $\delta$. The yellow lines indicate which particles of $\hat{p}_i$ have been matched to $\hat{\mu}$ to compute $\mathcal{W}_2(\hat{p}_i, \hat{\mu})$. (Right) In practice, \currot needs to rely on an approximation $\hat{\mathcal{V}}(\pi, \delta)$ of $\mathcal{V}(\pi, \delta)$, which is why a trust-region $d(\svec{c}_{p_i,n}, \svec{c}) {\leq} \epsilon$ is introduced to avoid the overly greedy exploitation of approximation errors. The indicated trust region (black dotted line) belongs to the non-opaque particle.}
    \label{fig:spherical:currot}
\end{figure}The parameter $\epsilon$ in (\ref{eq:spherical:currot-ind-opt}) limits the displacements of the particles within one update step, preventing the exploitation of faulty performance estimates $\hat{J}(\pi,\svec{c})$. The kernel bandwidth $h$ is set to a fraction of $\epsilon$, e.g., $h{=}0.3\epsilon$ in \cite{klink2022curriculum}, \change{given its purpose to capture the trend of $J(\pi, \svec{c})$} within the trust-region around $\svec{c}_{p_i,n}$. The $L$ contexts $\svec{c}_l$ \change{and episodic return $J_l$} used for predicting the agent performance are stored in two buffers, for whose update rules we refer to \cite{klink2022curriculum}. The $N$ target particles $\cvec{c}_{\mu, \change{\phi(n)}}$ are in each iteration sampled from $\mu(\svec{c})$ and the permutation $\change{\phi(n)}$ \change{assigning them to $\svec{c}_{p_i,n}$ is obtained by minimizing an assignment problem}
\begin{align}
    \mathcal{W}_2(\hat{p}_i, \hat{\mu}) = \min_{\change{\phi \in \text{Perm}(N)}} \left(\frac{1}{N} \sum_{n=1}^{N} d(\svec{c}_{p_i,n}, \svec{c}_{\mu, \change{\phi(n)}})^2 \right)^{\frac{1}{2}}. \label{eq:spherical:assignment}
\end{align}
If we can optimize $d(\svec{c}_{p_i,n}, \svec{c}_{\mu, \change{\phi(n)}})$ to zero for each particle in each iteration, we essentially sample from $\mu(\svec{c})$. \change{Figure \ref{fig:spherical:currot} shows a schematic visualization of \currot.} A crucial ingredient in the \currot algorithm is the distance function $d(\svec{c}_1, \svec{c}_2)$ that expresses the (dis)similarity between two learning tasks. So far, $d$ has been assumed to be the Euclidean distance in continuous spaces in \cite{klink2022curriculum}. A critical part of our experimental investigation of the benefit of curricula for learning tracking control will be the comparison of the Euclidean distance between the context vectors $\svec{c}_1$ and $\svec{c}_2$ and a Mahalanobis distance \cite{mahalanobis1936generalized}. In the following section, we describe this distance and other improvements that we benchmark in the experimental section.

\section{Improved Curriculum Generation} 

\noindent The \currot algorithm has so far been evaluated in rather low-dimensional scenarios, with two- or three-dimensional context spaces $\mathcal{C}$ that lend themselves to a Euclidean interpretation. In this section, we describe technical adjustments of the \currot algorithm that improve the creation of curricula over trajectories, i.e. over a high-dimensional context space $\mathcal{C}$ with a more intricate metric structure. 

\subsection{Affine Metrics} 
\label{sec:spherical:affine}

\noindent In \cite{klink2022curriculum}, the \currot algorithm has been evaluated under the assumption of a Euclidean metric 
\begin{align*}
    d(\svec{c}_1, \svec{c}_2) = \| \svec{c}_1 - \svec{c}_2 \|_2 = \sqrt{(\svec{c}_1 - \svec{c}_2)^T (\svec{c}_1 - \svec{c}_2)}    
\end{align*}
in continuous context spaces $\mathcal{C}$. For our trajectory representation, this corresponds to a Euclidean distance between elements in $\textrm{ker}(\cvec{\Psi}\change{(t_e)})$. However, according to Eq. (\ref{eq:spherical:closed-form-lti}), we know that the difference between two (one-dimensional) LTI system states is given by 
\begin{align*}
    \svec{x}_1(t) {-} \svec{x}_2(t) = \cvec{\Psi}(t) (\svec{u}_1 - \svec{u}_2).
\end{align*}
This observation allows us to compute the distance of the trajectories $\cvec{\gamma}_1(t)$, $\cvec{\gamma}_2(t)$ generated by $\svec{c}_1$, $\svec{c}_2$ via a Mahalanobis distance
\begin{align*}
    d_{\change{\cvec{\Psi}}}(\svec{c}_1, \svec{c}_2) &= \sqrt{(\svec{c}_1 - \svec{c}_2)^T \svec{A} (\svec{c}_1 - \svec{c}_2)}, \\ 
    \svec{A} &= \change{\cvec{\Gamma}_3^T \begin{bmatrix} \cvec{\Psi}_3(t_s) \\
    \cvec{\Psi}_3(t_1) \\
    \vdots \\
    \cvec{\Psi}_3(t_e) \end{bmatrix}^T \begin{bmatrix} \cvec{\Psi}_3(t_s) \\
    \cvec{\Psi}_3(t_1) \\
    \vdots \\
    \cvec{\Psi}_3(t_e) \end{bmatrix} \cvec{\Gamma}_3}
\end{align*}
\change{where $\cvec{\Gamma} {\in} \mathbb{R}^{K \times {K-3}}$ maps the elements $\tilde{\svec{u}} {\in} \text{ker}(\cvec{\Psi}(t_e))$ to jerk sequences $\svec{u}$. We then ``repeat'' $\cvec{\Gamma}$ and $\cvec{\Psi}(t)$ to capture the three spatial dimensions, forming the block diagonal matrices $\cvec{\Gamma}_3 {=} \text{blkdiag}\left(\{\cvec{\Gamma}\}_{n=1}^3\right)$ and $\cvec{\Psi}_3(t) {=} \text{blkdiag}\left(\{\cvec{\Psi}(t)\}_{n=1}^3\right)$.} The Mahalanobis distance can be computed with no change to the algorithm by whitening the contexts $\svec{c}$ and computing the Euclidean distance in the whitened space.

\begin{figure}
    \centering
    \includegraphics{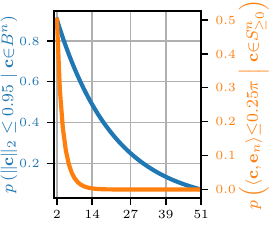} \includegraphics{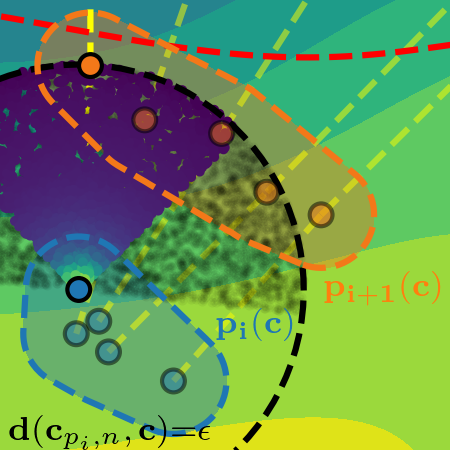}
    \caption{(Left) In higher dimensions, the Euclidean norm of a vector $\textcolor{c0blue}{\|\svec{c}\|_2}$ in the $n$-dimensional ball $\textcolor{c0blue}{B^n}$ increasingly converges to one. The angle $\textcolor{c1orange}{\change{\angle \svec{c} \svec{e}_n}}$ between a context $\textcolor{c1orange}{\svec{c}}$ in the $n$-dimensional half sphere $\textcolor{c1orange}{S^n_{\geq 0}}$ and any $n$-dimensional vector $\textcolor{c1orange}{\svec{e}_n}$ is with increasing certainty larger than $\textcolor{c1orange}{45^{\circ} {=} 0.25 \pi}$. (Right) This behavior requires adapting the sampling-based optimization of Objective (\ref{eq:spherical:currot-ind-opt}) to sample those unit vectors that make an angle of less than $45^{\circ}$ with a descent direction and scale them uniformly in $[0,\epsilon]$. Unlike the default \currot sampling scheme (black samples), this sampling scheme (colored dots, color indicates density) more robustly finds descent directions in high-dimensional tasks.}
    \label{fig:spherical:currot_sampling}
\end{figure}

\subsection{Sampling-Based Optimization}
\noindent The optimization of (\ref{eq:spherical:currot-ind-opt}) is carried out in parallel by uniformly sampling contexts in an $n$-dimensional $\epsilon$-half ball
\begin{align*}
    B^n_{\geq 0}(\svec{c}_{p_i,n}, \epsilon) = \big\{ \svec{c} \ \big| \  &\| \svec{c} - \svec{c}_{p_i,n} \|_2 \leq \epsilon \ \land \\
    & \langle \svec{c} - \svec{c}_{p_i,n}, \svec{c}_{\mu, \change{\phi(n)}} \change{- \svec{c}_{p_i,n}} \rangle \geq 0 \big\}
\end{align*}
around $\svec{c}_{p_i,n}$ and selecting the sample with minimum distance to $\svec{c}_{\mu, \change{\phi(n)}}$ that fulfills the performance constraint. \change{$\langle \cdot, \cdot \rangle$ denotes the dot product.} In higher dimensions, this sampling scheme faces two problems. Firstly, the mass of a ball is increasingly concentrated on the surface for higher dimensions, resulting in samples that are increasingly concentrated at the border of the trust region. Secondly, the chance of sampling a context $\svec{c}$ for which $d(\svec{c}, \svec{c}_{\mu,\change{\phi(n)}}) < d(\svec{c}, \svec{c}_{p_i,n})$ decreases dramatically for higher dimensions as soon as $\| d(\svec{c}_{p_i,n}, \svec{c}_{\mu,\change{\phi(n)}}) \| {\leq} \epsilon$. To remedy both problems, we first sample unit vectors that make an angle less than $\theta\change{=0.25\pi}$ with the descent direction $\svec{c}_{\mu,\change{\phi(n)}} {-} \svec{c}_{p_i,n}$. Such unit vectors can be sampled using, e.g., the sampling scheme described in \cite{asudeh2018obtaining}. We then scale these search direction vectors by a scalar that we uniformly sample from the interval $[0, \epsilon]$. Figure \ref{fig:spherical:currot_sampling} contrasts the new sampling scheme with the one introduced with the \currot algorithm in \cite{klink2022curriculum}.

\subsection{Tracking Metrics other than Reward}

\noindent The constraint $p(\mathcal{V}(\pi, \delta)) {=} 1$ in Objective (\ref{eq:spherical:currot}) is controlling the curriculum's progression towards $\mu(\svec{c})$ by preventing it from sampling contexts in which the agent does not fulfill a performance threshold $\delta$. We generalize this constraint to define $\mathcal{V}$ based on an arbitrary function $M(\pi, \svec{c}) {\in} \mathbb{R}$ obtained from a rollout of the policy $\pi$ in a context $\svec{c}$. We hence define $\mathcal{V}(\pi, \delta) = \left\{ \svec{c} \in \mathcal{C} \middle| M(\pi, \svec{c}) \geq \delta \right\}$.
The same Nadaraya-Watson kernel regression introduced in Section \ref{sec:spherical:currot} can approximate $M(\pi, \svec{c})$. In our setting, the increased flexibility enables restricting training to those trajectories for which the agent can stabilize the pendulum throughout almost the whole episode, i.e., almost all of the $1500$ episode steps. Encoding this restriction via a fixed lower bound on the episode return is hard to achieve due to, e.g., regularizing terms on the joint velocities and the penalty for non-precise tracking of $\cvec{\gamma}(t)$. \change{These terms can result in highly differing returns for episodes in which the agent stabilized the pendulum the entire episode}.

\subsection{GPU Implementation}

\noindent The authors of \cite{klink2022curriculum} provide an implementation of \currot in NumPy \cite{harris2020array} and SciPy \cite{virtanen2020scipy}, computing the assignment to the target distribution particles using the SciPy-provided linear sum assignment solver. Given the large number of parallel simulations \change{that we utilize}, our application of \currot needed to work with a large number of particles $N$ and contexts for performance prediction $L$. We hence created a GPU-based implementation using PyTorch \cite{paszke2019pytorch}. To solve the assignment problem (\ref{eq:spherical:assignment}), we implemented a default auction algorithm \cite{bertsekas2009auction} using the PyKeOps library \cite{charlier2021jmlr}, which provides highly efficient CUDA routines for reduction operations on large arrays. We also use the PyKeOps library for the Nadaraya-Watson kernel regression. The GPU implementation of \currot and the code for running the experiments described in the next section will be made publicly available upon acceptance.

\section{Experiments}

\begin{figure}[t]
    \centering
    \includegraphics{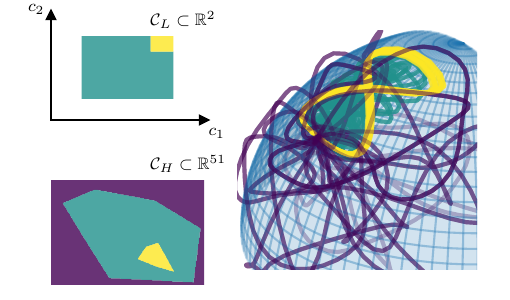}
    \caption{A visualization of the eight-shaped target trajectories $\cvec{\gamma}(t)$ (in yellow) that the learning agent is required to track in our experiments. The trajectories are projected onto a dome that is centered around the robot. Due to the particular shape of the trajectories, we can represent them via both a low- and high-dimensional parametric description, providing the possibility to test how \currot scales to high-dimensional context representations. Compared to the low-dimensional representation, eight-shaped trajectories are only a small part of the full, high-dimensional context space.}
    \label{fig:spherical:eight_task}
\end{figure}

\begin{figure*}[t]
    \centering
    \includegraphics{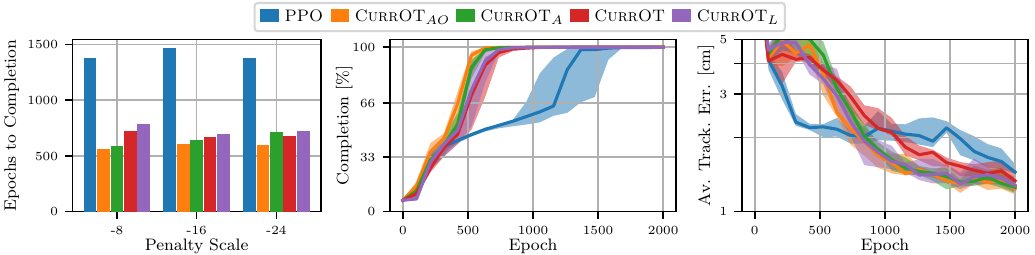}
    \caption{(Left) Ablation over different tipping penalties $\alpha$ and their effect on the average required number of epochs until successfully completing the target trajectories. (Middle) Completion rate (i.e. fraction of maximum steps per episode) over epochs for $\alpha{=}-8$ for different learning methods. (Right) Achieved tracking error during the agent lifetime over epochs for different learning methods. In the middle and right plot, thick lines represent the median, and the shaded areas represent interquartile ranges. Statistics are computed from $10$ seeds.}
    \label{fig:spherical:eight_performance}
\end{figure*}

\noindent In this section, we answer the following questions by evaluating the described learning system in simulation as well as on the real system:
\begin{itemize}
    \item Do curricula stabilize or speed up learning in \change{the trajectory tracking} task?
    \item How do the proposed changes to the \currot algorithm alter the generated curriculum and its benefit on the learning agent?
    \item Does the behavior learned in simulation transfer to the real system?
\end{itemize}
The experiment requires the agent to track eight-shaped trajectories projected onto a sphere (Figure \ref{fig:spherical:eight_task}). The target distribution $\mu(\cvec{\gamma})$ of tasks encodes eight-shaped trajectories whose maximal distance to the starting position is $0.36$-$0.4$m in the $x$-dimension and $0.18$-$0.2$m in the $y$-dimension. We choose the $z$-coordinate of the trajectory such that the trajectory has a constant distance to the first joint of the Barrett WAM, i.e., moves on a sphere centered on this joint. \\
We chose this particular task since the trajectories encoded by the target distributions $\mu(\cvec{\gamma})$ can, in addition to the parameterization via jerks, be parameterized in a two-dimensional parameter space, which enables us to benchmark how the \currot algorithm proposed in \cite{klink2022curriculum} behaves both in low- and high-dimensional task parameterizations. \\
For the two-dimensional representation, we represent the maximum distance in $x$- and $y$-dimension when generating curricula in the two-dimensional context space $\mathcal{C}_{L} {\subset} \mathbb{R}^2$. When representing trajectories via jerks, we compose the jerk sequence $\svec{u}$ of $K{=}20$ constant segments evenly spread in the interval $[1,10.5]$. The first- and last second of each trajectory is always stationary at $\svec{x}(t_s)$. Hence, the actual movement happens within $[1,11]$ seconds. Due to constraint (\ref{eq:spherical:closing-constraint}) of starting and ending in $\svec{x}(t_s)$, the parameterization reduces to $17$ dimensions for each task space dimension, i.e., $\mathcal{C}_{H} {\subset} \mathbb{R}^{51}$. For building the curricula, we define the set $\mathcal{V}(\pi, \delta)$ to contain those trajectories for which the policy manages to keep the pendulum upright for at least $1400$ steps, i.e., those trajectories which fully complete their movements during the lifetime of the agent (remember that the policy is stationary for the last second, i.e., the last $125$ out of $1500$ steps). \\
We ablate the default \ppo learner as well as four ablations of the \currot method introduced in Section \ref{sec:spherical:currot}
\begin{itemize}
    \item \currot: The default algorithm as introduced by \cite{klink2022curriculum} using our GPU-based implementation and using $M(\pi, \svec{c})$ instead of $J(\pi, \svec{c})$ to define $\mathcal{V}(\pi, \delta)$.
    \item \lowdimcurrot: The default algorithm exploiting the low-dimensional parameterization of the target trajectories to generate curricula in $\mathbb{R}^2$ instead of $\mathbb{R}^{51}$.
    \item \affinecurrot: A variation of \currot that uses the metric $d_{\change{\cvec{\Psi}}}$ to capture the dissimilarity between the generated trajectories rather than the context variables.
    \item \newcurrot: The version of \currot that combines the use of $d_{\change{\cvec{\Psi}}}$ with improvements to the sampling-based optimization of Objective (\ref{eq:spherical:currot-ind-opt}).
\end{itemize}
For all curricula, we choose the trust-region parameter $\epsilon$ of Objective (\ref{eq:spherical:currot-ind-opt}) according to method described in \cite{klink2022curriculum}, i.e. setting $\epsilon {\approx} 0.05 \max_{\svec{c}_1, \svec{c}_2 \in \mathcal{C}} d(\svec{c}_1, \svec{c}2)$. \change{All curricula train on an initial distribution $p_0(\svec{c})$ of trajectories that barely deviate from the initial position} until $\mathbb{E}_{p_0} \left[ \hat{M}(\pi, \svec{c}) \right] {\geq} \delta$, at which point the methods start updating the context distribution. All methods train for $262$ million learning steps, where a policy update is performed after $64$ environment steps, resulting in $64 \cdot 2048 {=} 131072$ samples generated between a policy update.

\subsection{Quantitative Results}

\noindent Figure \ref{fig:spherical:eight_performance} shows the performance of the learned policies. More precisely, we show the average tracking error during the agent's lifetime and the number of completed trajectory steps on $\mu(\cvec{\gamma})$. While the tracking errors behave similarly between the investigated methods, the curricula shorten the required training iterations until we can track complete target trajectories. \\
The results indicate that by first focusing on trajectories that can be tracked entirely and then gradually transforming them into more complicated ones, we avoid wasteful biased sampling of initial parts of the trajectory due to system resets once the pendulum falls over. \\
We additionally ablate the results over the penalty term $\alpha$ that the agent receives when the pendulum topples over (Eq. \ref{eq:spherical:reward}). Figure \ref{fig:spherical:eight_performance} shows that its influence on the learning speed of the agent is limited, as the epochs required by \ppo to track the target trajectories completely stays relatively constant even when increasing $\alpha$ by a factor of three. \\
For the curriculum methods themselves, we can make two observations. First, in the high-dimensional context space $\mathcal{C}_{H} \subset \mathbb{R}^{51}$, all curricula learn the task reliably with comparable learning speed. Second, operating in the low-dimensional context space $\mathcal{C}_{L} \subset \mathbb{R}^2$ does not lead to faster learning. Both results surprised us since, in the high-dimensional context space, we expected that exploiting the structure of the context space via $d_{\change{\cvec{\Psi}}}(\svec{c}_1, \svec{c}_2)$ and the improved optimization would significantly improve the curricula. Furthermore, we expected the low-dimensional representation $\mathcal{C}_{L} \subset \mathbb{R}^2$ to ease the performance estimation $\hat{J}(\pi, \svec{c})$ via kernel regression since we can more densely populate the context space $\mathcal{C}_{L}$ with samples of the current agent performance. The following section highlights why the observed performance did not behave according to our expectations.

\subsection{Qualitative Analysis of Generated Curricula}

\begin{figure*}[t]
    \centering
    \begin{subfigure}[b]{0.24\textwidth}
        \centering
        \includegraphics{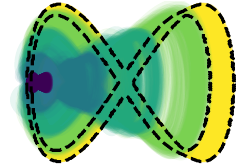}
        \caption{\newcurrot}
    \end{subfigure}
    \begin{subfigure}[b]{0.24\textwidth}
        \centering
        \includegraphics{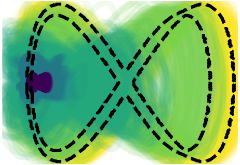}
        \caption{\affinecurrot}
    \end{subfigure}
    \begin{subfigure}[b]{0.24\textwidth}
        \centering
        \includegraphics{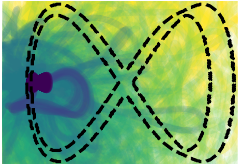}
        \caption{\currot}
    \end{subfigure}
    \begin{subfigure}[b]{0.24\textwidth}
        \centering
        \includegraphics{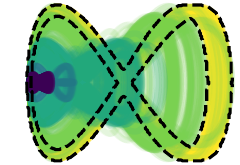}
        \caption{\lowdimcurrot}
    \end{subfigure}
    \caption{Evaluation of training distributions $p_i$ for different ablations of \currot. Brighter colors indicate later iterations. The black dotted line indicates the "boundaries" of the support of the target distribution. Note that the distributions have been projected onto a 2D plane, omitting the z-coordinate.}
    \label{fig:spherical:eight_trajectory_evolution}
\end{figure*}

\begin{figure}[t]
    \centering
    \includegraphics{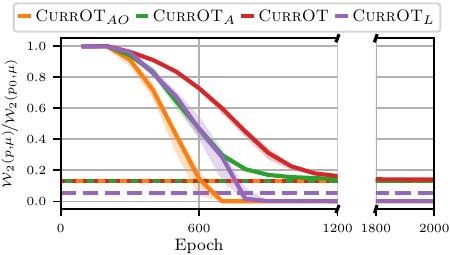}
    \caption{Evolution of Wasserstein distance $\mathcal{W}_2(p, \mu)$ compared to the initial distance over learning epochs for different variants of \currot. The dashed horizontal lines represent $\frac{\epsilon}{\mathcal{W}_2(p_0, \mu)}$, i.e., the fraction between the trust-region $\epsilon$ for Objective (\ref{eq:spherical:currot-ind-opt}) and the initial Wasserstein distance $\mathcal{W}_2(p_0, \mu)$. Thick lines represent the medians, and shaded areas visualize interquartile ranges. Statistics are computed from $10$ seeds.}
    \label{fig:spherical:eight_wasserstein_evolution}
\end{figure}

\noindent To better understand the dynamics of the generated curricula, we visualize the generated trajectories throughout different learning epochs and the evolution of the Wasserstein distance $\mathcal{W}_2(p_i, \mu)$ in Figures \ref{fig:spherical:eight_trajectory_evolution} and \ref{fig:spherical:eight_wasserstein_evolution}. Focusing on the evolution of Wasserstein distances shown in Figure \ref{fig:spherical:eight_wasserstein_evolution}, we can see that only \newcurrot and \lowdimcurrot can converge to the target distribution, achieving zero Wasserstein distance. \currot and \affinecurrot do not converge to $\mu(\svec{c})$ after initially exhibiting fast progression towards $\mu(\svec{c})$ but slowing down as the Wasserstein distance approaches the value of the trust region parameter $\epsilon$. This slowing-down behavior is precisely due to the naive sampling in the half-ball of the default \currot algorithm, which we discussed in Section \ref{sec:spherical:currot}. If the target contexts are well outside the trust region, even samples that do make an angle larger than $0.25\pi$ with the descent direction $\svec{c}_{\mu, \change{\phi(n)}} {-} \svec{c}_{p_i,n}$ decrease the distance to the target $\svec{c}_{\mu, \change{\phi(n)}}$. Once the target contexts are on or within the boundary of the trust region, the effect visualized in Figure \ref{fig:spherical:currot_sampling} takes place, preventing further approach to the target samples. As shown in Figure \ref{fig:spherical:eight_trajectory_evolution}, the effect of the resulting bias on the generated trajectories greatly depends on the chosen metric. By measuring dissimilarity via the Euclidean distance between contexts $\svec{c}_1$ and $\svec{c}_2$, \currot generates trajectories that behave entirely differently from the target trajectories during the initial and later stages of training. Incorporating domain knowledge via $d_{\change{\cvec{\Psi}}}$ allows \affinecurrot to generate trajectories with similar qualitative behavior to the target trajectories throughout the learning process. \\
While underlining the importance of the proposed improvements in \newcurrot, the high performance achieved by \currot and \affinecurrot, despite the potentially strong dissimilarity in generated trajectories, stresses a critical observation: The success of a curriculum is inherently dependent on the generalization capability of the learning agent. By conditioning the policy behavior on limited-time lookahead windows of the target trajectory $\svec{T}_t$, the learning agent seems capable of generalizing well to unseen trajectories as long as those trajectories visit similar task-space positions as the trajectories in the training distribution. Consequently, the failure of \currot to generate trajectories of similar shape to those in $\mu(\cvec{\gamma})$ is compensated for by the generalization capabilities of the learning agent. \change{All in all, the results indicate that convergence of $p_i(\svec{c})$ to $\mu(\svec{c})$ is only a sufficient condition for good agent performance on $\mu(\svec{c})$, but not a necessary one.}

\subsection{Alternative Trajectory Representation}
\label{sec:spherical:alternative}

\begin{figure}
    \centering
    \includegraphics{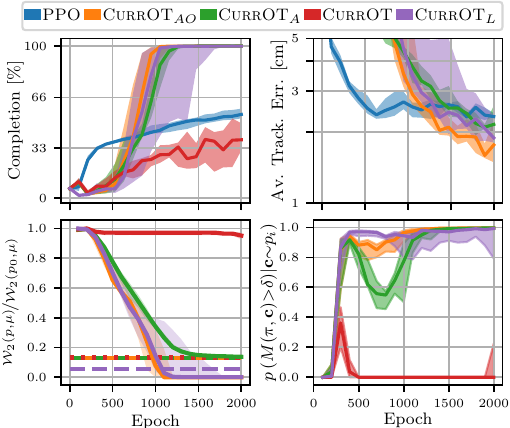}
    \caption{(Top Left) Completion rate (i.e. fraction of maximum steps per episode) over epochs for $\alpha{=}-8$ for different learning methods. (Top Right) Tracking error achieved during the agent lifetime over epochs for different learning methods. (Bottom Left) Wasserstein distance between training- and target distribution over epochs. (Bottom Right) Percentage that $M(\pi, \svec{c}) \geq \delta$ on the training distribution $p_i(\svec{c})$ over epochs. We show median and interquartile ranges that are computed from $10$ seeds.}
    \label{fig:spherical:eight_performance_ablation}
\end{figure}

\noindent The surprising effectiveness of \currot, despite its ignorance of the context space structure, led us to conclude that the policy structure leads to rather strong generalization capabilities of the agent, concealing shortcomings of the generated curricula. We changed the policy architecture to test this hypothesis, replacing the trajectory lookahead $\svec{T}_t$ simply by the contextual parameter $\svec{c} \in \mathbb{R}^{51}$ and the current time index $t \in \mathbb{R}$. While this representation still contains all required information about the desired target position $\cvec{\gamma}(t)$ at time step $t$, it does not straightforwardly allow the agent to exploit common subsections of two different trajectories. Figure \ref{fig:spherical:eight_performance_ablation} visualizes the results of this experiment. Comparing Figures \ref{fig:spherical:eight_performance} and \ref{fig:spherical:eight_performance_ablation}, we see that the different context representation slows down the learning progress of all curricula and consequently leads to higher tracking errors after $2000$ epochs. We also see that the lookahead $\svec{T}_t$ benefits learning with \ppo, as with the new context representation, none of the $10$ seeds learn to complete the trajectory within $2000$ epochs. More importantly, we see how the inadequate metric of \currot now leads to a failure in the curriculum generation, with $\mathcal{W}_2(p_i, \mu)$ staying almost constant for the entire $2000$ epochs as the agent struggles to solve the tasks in the curriculum. This failure to generate tasks of adequate complexity is also shown in Figure \ref{fig:spherical:eight_performance_ablation}, where we visualize the percentage of tasks $\svec{c}$ sampled by the curriculum for which $M(\pi, \svec{c}) \geq \delta$. As we can see, this percentage drops to zero under \currot once the algorithm starts updating $p_0(\svec{c})$. The other curricula maintain a non-zero success percentage. We can also observe a pronounced drop in success rate for \affinecurrot, which is not present for \newcurrot, potentially due to the more targeted sampling in the approximate update step of the context distribution resulting in more similar trajectories. \\
Apart from this ablation, we also performed experiments for increasing context space dimensions, generating curricula in up to $399$-dimensional context spaces. The results in Appendix \ref{app:spherical:high_dim} show that \newcurrot generates beneficial curricula across all investigated dimensions and trajectory representations, while the curricula of \affinecurrot become less effective in higher dimensions for the alternative trajectory representation presented in this section. For \currot, the resulting picture stays unchanged with poor observed performance for the alternative trajectory representation regardless of the context space dimension.

\subsection{Real Robot Results}

\begin{figure}[t]
    \centering
    \begin{subfigure}[b]{0.15\textwidth}
        \centering
        \includegraphics{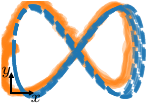}
        \includegraphics{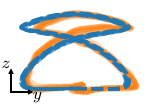}
        \caption{\ddp}
    \end{subfigure}
    \begin{subfigure}[b]{0.15\textwidth}
        \centering
        \includegraphics{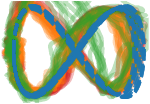}
        \includegraphics{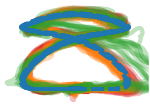}
        \caption{\ppo}
    \end{subfigure}
    \begin{subfigure}[b]{0.15\textwidth}
        \centering
        \includegraphics{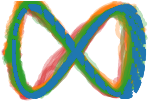}
        \includegraphics{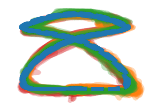}
        \caption{\newcurrot}
    \end{subfigure}
    \caption{Generated trajectories on the real robot. We visualize projections to the $xy$- (top) and $yz$-plane (bottom) to highlight the three-dimensional nature of the trajectory. The reference trajectories are shown in blue. Other colors indicate trajectories that have been generated by the different (learned) controllers. For \ppo and \newcurrot, we evaluate the three best-performing seeds, indicated by colors.}
    \label{fig:spherical:real_robot_trajectories}
\end{figure}

\begin{figure*}[t]
    \centering
    \begin{subfigure}[b]{0.19\textwidth}
        \centering
        \includegraphics[width=\textwidth,trim=0 100 0 0, clip]{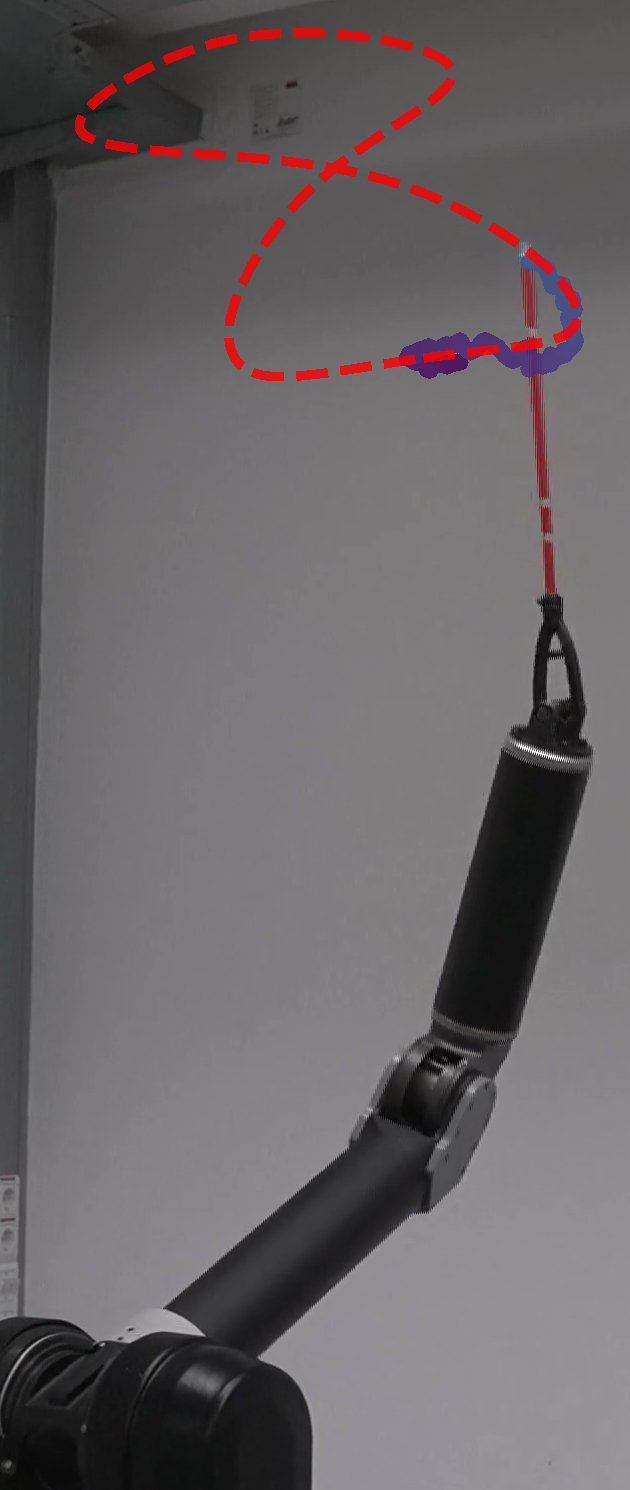}
        \caption{$t\approx4.0$}
    \end{subfigure}
    \begin{subfigure}[b]{0.19\textwidth}
        \centering
        \includegraphics[width=\textwidth,trim=0 100 0 0, clip]{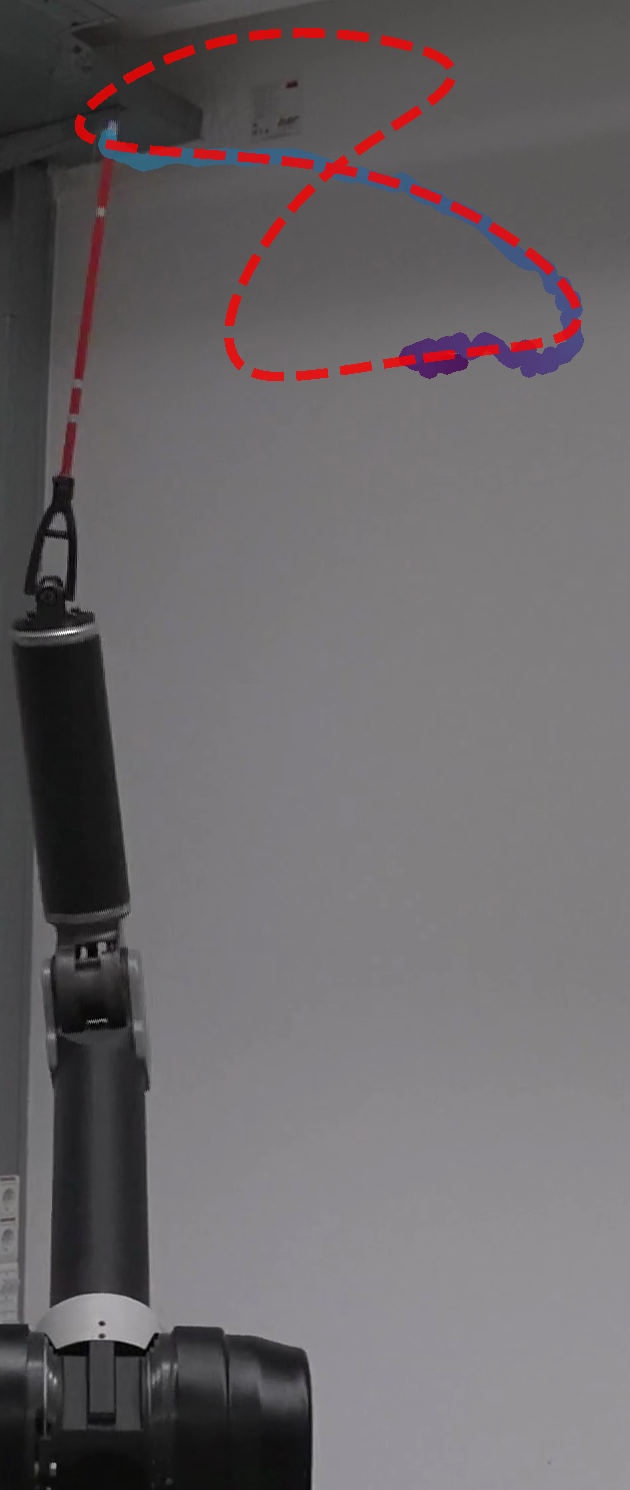}
        \caption{$t\approx5.2$}
    \end{subfigure}
    \begin{subfigure}[b]{0.19\textwidth}
        \centering
        \includegraphics[width=\textwidth,trim=0 100 0 0, clip]{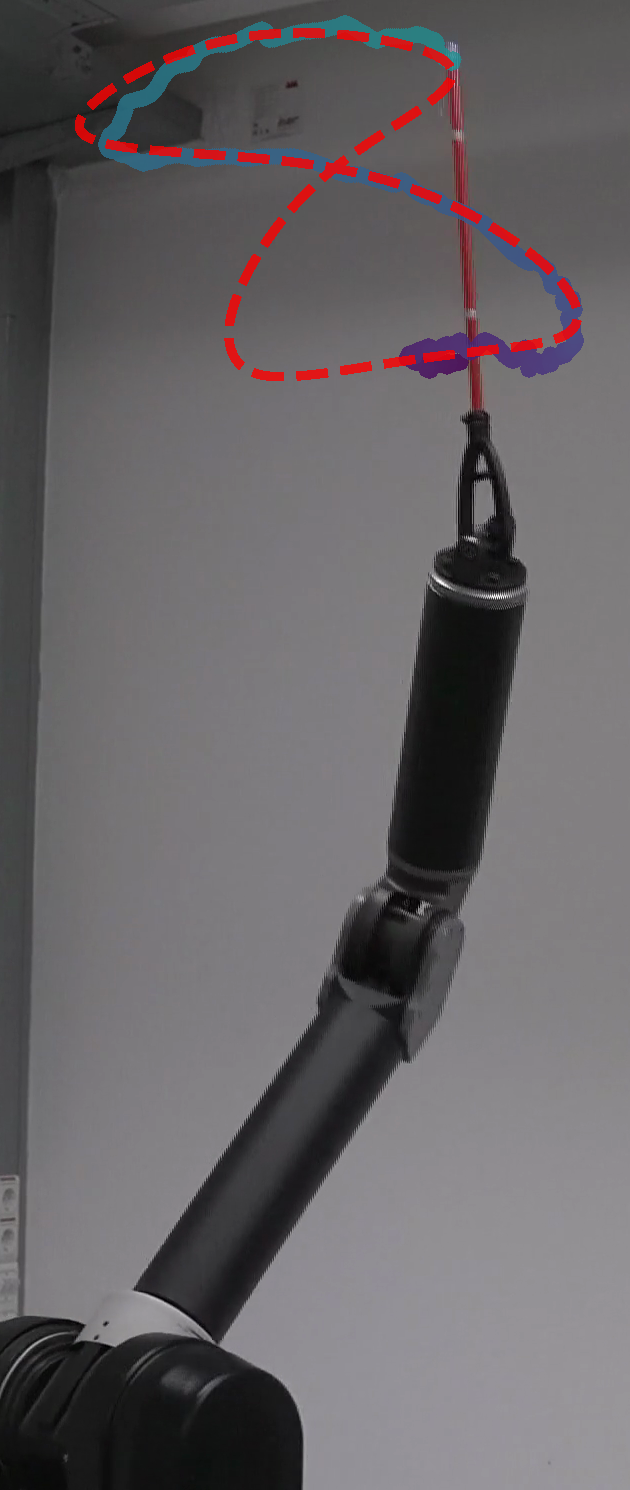}
        \caption{$t\approx6.5$}
    \end{subfigure}
    \begin{subfigure}[b]{0.19\textwidth}
        \centering
        \includegraphics[width=\textwidth,trim=0 100 0 0, clip]{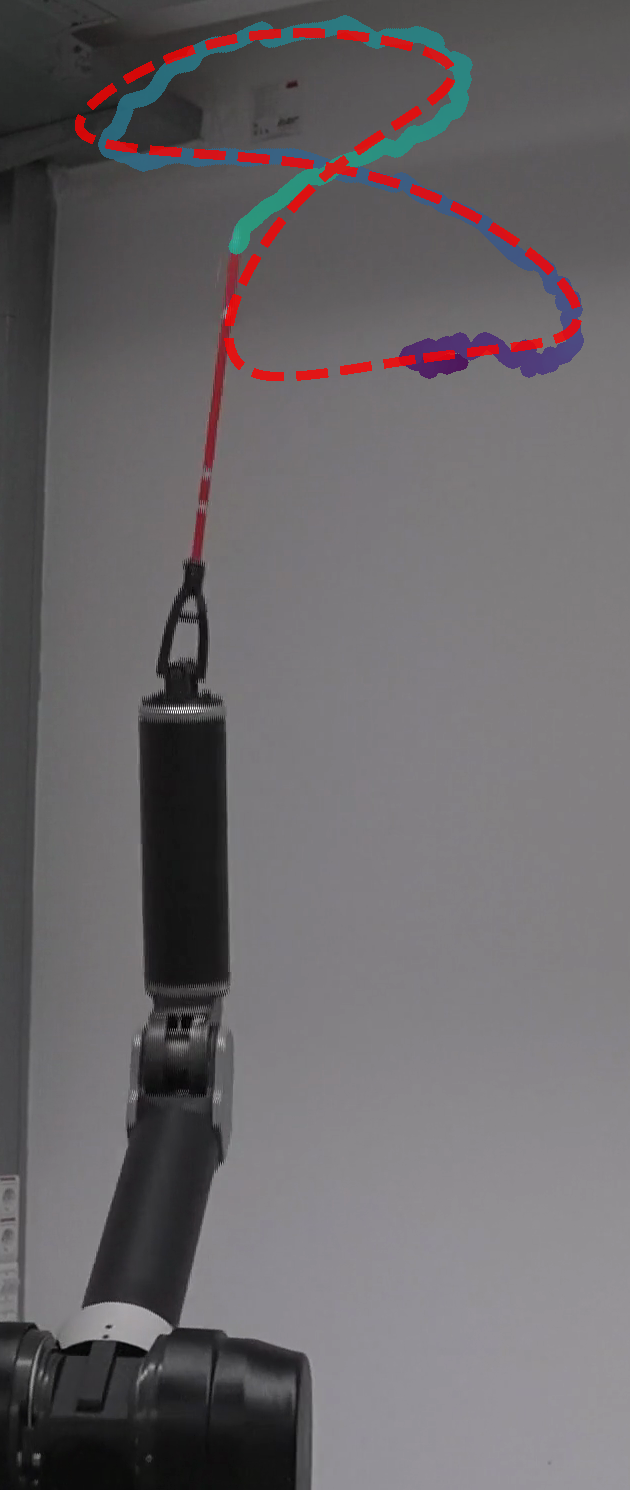}
        \caption{$t\approx8.0$}
    \end{subfigure}
    \begin{subfigure}[b]{0.19\textwidth}
        \centering
        \includegraphics[width=\textwidth,trim=0 100 0 0, clip]{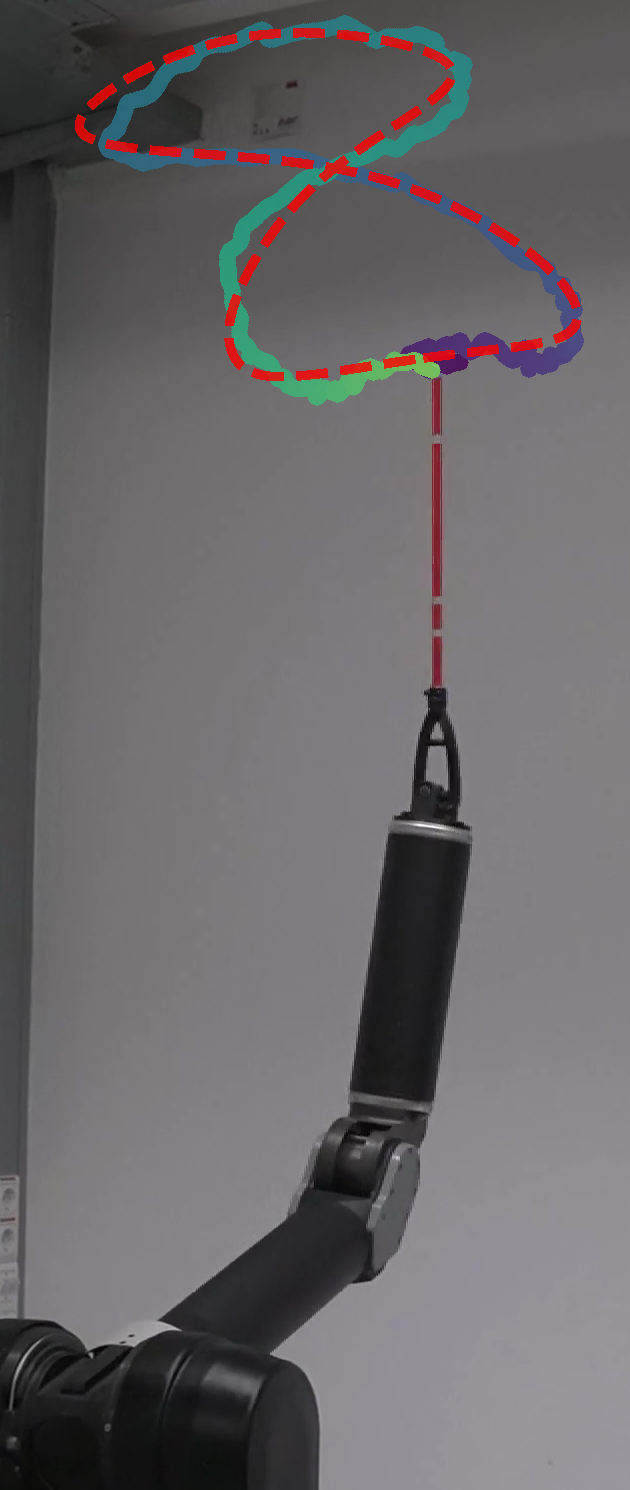}
        \caption{$t\approx11$}
    \end{subfigure}
    \caption{Snapshots of the policy learned with \newcurrot during execution on the real robot. The dotted red line visualizes the target trajectory to be tracked by the policy. The generated trajectory is visualized by the colored line, where brighter colors indicate later time-steps.}
    \label{fig:spherical:real_robot_execution}
\end{figure*}

\noindent To assess transferability to the real world and put the achieved results into perspective, we evaluated the three best policies learned with \ppo and \newcurrot for $\alpha {=} -8$ on the real robot and compared them to an optimal control baseline. We evaluated each seed on $10$ trajectories sampled from $\mu(\cvec{\gamma})$. The target trajectories are shown in Figure \ref{fig:spherical:real_robot_trajectories} and Figure \ref{fig:spherical:real_robot_execution} shows snapshots of the policy execution on the real system. Given the architectural simplicity of the agent policy, it was easy to embed it in a C++-based ROS \cite{Quigley09} controller using the Eigen library, receiving the pole information from Optitrack via UDP packets. The execution time of the policy network was less than a millisecond and hence posed no issue for our target control frequency of $125$ Hz. Given that the pole starts in an upright position during training, we attached a thread to the tip of the pendulum to stabilize the pendulum via a pulley system before starting the controller. We then simultaneously release the thread and start the controller. Given the negligible weight of the thread, we did not observe any interference with the pole. To ensure safety during the policy execution, we first executed the policies in a MuJoCo simulation embedded in the ROS ecosystem. We monitored the resulting minimum- and maximum joint positions $\svec{q}_{\text{min}}$ and $\svec{q}_{\text{max}}$, and defined a safe region $\mathcal{S}$, which the agent is not supposed to leave during execution on the real system
\begin{align*}
    \mathcal{S} = [\bar{\svec{q}} - 1.25 (\svec{q}_{\text{max}} - \svec{q}_{\text{min}}), \bar{\svec{q}} + 1.25 (\svec{q}_{\text{max}} - \svec{q}_{\text{min}})],
\end{align*}
where $\bar{\svec{q}} = \nicefrac{1}{2} (\svec{q}_{\text{min}} + \svec{q}_{\text{max}})$. Inspired by the results of~\cite{vu2021fast}, we decided to compare the results of \ppo and \newcurrot to an optimal control baseline that we obtained by computing a time-varying linear feedback controller using the differential dynamic programming (\ddp) algorithm implemented in the Crocoddyl library \cite{mastalli20crocoddyl}. At the convergence of \ddp, we can obtain a time-varying linear controller from the internally computed linearization of the dynamics on the optimal trajectory. We use the same cost function as for the reinforcement learning agent, simply removing the penalty term for tipping the pendulum, as the gradient-based \ddp does not run into danger of tipping the pendulum. The obtained time-varying controller requires access to full state information, i.e., position and velocity of the robot and pole, which we infer using a high-gain non-linear observer \cite{khalil2014high}, whose gains we tuned on the real system using the synthesized controller to achieve the best tracking performance. \begin{table}[b!]
\renewcommand{\arraystretch}{1.25}
\caption{Mean and standard deviation of tracking errors achieved with different controllers. We evaluate both on a ROS-embedded MuJoCo simulation (Sim) and the real robot (Real). For \ppo and \newcurrot, the color of the seeds corresponds to the trajectories shown in Figure \ref{fig:spherical:real_robot_trajectories}. In each row, statistics are computed from $20$ policy executions.}
\label{tab:spherical:real_robot_results}
\centering
\vspace{-15pt}
\begin{tabular}{c|c|c|c|c}
\multicolumn{5}{c}{} \\
\multicolumn{5}{c}{\ddp} \\
\hline
Gain & \multicolumn{2}{c|}{Sim} & \multicolumn{2}{c}{Real} \\
\hline 
& Completion & Error [cm] & Completion & Error [cm] \\
\hline
High & $1.00$ & $0.75 {\pm} 0.02$ & - & -  \\
Low & $1.00$ & $2.58 {\pm} 0.11$ & $1.00$ & $3.11 {\pm} 0.12$ \\
\multicolumn{5}{c}{} \\[-1em]
\multicolumn{5}{c}{\newcurrot} \\
\hline
Seed & \multicolumn{2}{c|}{Sim} & \multicolumn{2}{c}{Real} \\
\hline 
& Completion & Error [cm] & Completion & Error [cm] \\
\hline
\textcolor{c3red}{2} & $1.00$ & $1.98 {\pm} 0.06$ & $1.00$ & $2.30 {\pm} 0.16$ \\
\hline
\textcolor{c1orange}{4} & $1.00$ & $1.80 {\pm} 0.07$ & $1.00$ & $2.44 {\pm} 0.09$ \\
\hline
\textcolor{c2green}{5} & $1.00$ & $1.85 {\pm} 0.07$ & $1.00$ & $2.10 {\pm} 0.13$ \\
\hline \hline
Avg. & $1.00$ & $1.88 {\pm} 0.10$ & $1.00$ & $2.28 {\pm} 0.19$ \\
\multicolumn{5}{c}{} \\[-1em]
\multicolumn{5}{c}{\ppo} \\
\hline
Seed & \multicolumn{2}{c|}{Sim} & \multicolumn{2}{c}{Real} \\
\hline 
& Completion & Error [cm] & Completion & Error [cm] \\
\hline
\textcolor{c3red}{1} & $1.00$ & $3.10 {\pm} 0.14$ & $1.00$ & $2.82 {\pm} 0.20$ \\
\hline
\textcolor{c1orange}{5} & $1.00$ & $2.30 {\pm} 0.08$ & $1.00$ & $2.71 {\pm} 0.20$ \\
\hline
\textcolor{c2green}{8} & $1.00$ & $2.15 {\pm} 0.07$ & $0.55$ & $4.18 {\pm} 0.66$ \\
\hline \hline
Avg. & $1.00$ & $2.52 {\pm} 0.43$ & $0.85$ & $3.07 {\pm} 0.68$ \\
\end{tabular}
\end{table}\\
We performed the tracking experiments twice on different days, obtaining $20$ trajectories per seed and method, from which we can compute statistics. Figure \ref{fig:spherical:real_robot_trajectories} visualizes the result of the policy rollouts on the real system, and Table \ref{tab:spherical:real_robot_results} provides quantitative data. As shown in Figure \ref{fig:spherical:real_robot_trajectories}, the policies learned with \newcurrot seem to track the reference trajectories more precisely than the other methods. This impression is backed up by the data in Table \ref{tab:spherical:real_robot_results}, where the average tracking performance of both \ddp and \ppo on the real robot is about $35\%$ worse than that of \newcurrot. Comparing the results of the ROS-embedded MuJoCo simulation and the execution on the real robot, we see that, on average, the performance on the real system is about $20\%$ worse across all methods. Regarding reliability, one out of the three policies learned with \ppo did not reliably perform the tracking task, as it left the safe region $\mathcal{S}$ during execution. We can observe a significantly worse real robot tracking performance for this particular seed in Figure \ref{fig:spherical:real_robot_trajectories}. Looking at the \ddp results again, we see a distinction between high and low gains. The high gain setting corresponds to precisely using reward function \ref{eq:spherical:reward}, resulting in the best tracking performance across all methods in simulation. However, the high gains of the generated time-varying linear feedback controllers resulted in unstable behavior in the real system. To obtain stable controllers in the real system, we needed to increase the regularization of actions, position, and velocity by a factor of $30$. We assume that more sophisticated methods that better account for uncertainty in the model parameters could further improve performance. Since our baseline only aims to put the learned behavior into perspective, we did not explore such advanced methods. Instead, we interpreted the results as evidence that deep RL-based methods can learn precise control of highly unstable systems comparable to classical control methods. \\
Looking back at Figure \ref{fig:spherical:eight_performance}, we see that the tracking errors in the ROS-embedded MuJoCo simulation in Table \ref{tab:spherical:real_robot_results} are slightly worse than the error observed in Isaac Sim, where \ppo consistently achieved a tracking error of less than $2$cm, and \newcurrot consistently achieved a tracking error of less than $1.5$cm. This performance gap may be caused by our approximate modeling of actuation delay and tendons, and we expect additional efforts on system identification, modeling, and domain randomization to close this gap.

\section{Conclusions}

\noindent We presented an approach that learns a tracking controller for an inverted spherical pendulum mounted to a four degrees-of-freedom Barret Whole Arm Manipulator. We showed that increasingly available massively parallel simulators allow off-the-shelf reinforcement learning algorithms paired with curricula to reliably learn this non-trivial partially observable control task across policy- and task-space representations. Our evaluations of curricula and their effect on learning success showed multiple interesting results. Apart from confirming the sample-complexity benefit of learning the tracking task via curricula, we showed that a) the generation of curricula is possible in high-dimensional context spaces and b) that high-dimensionality does not need to make the curriculum generation less efficient. However, we also saw that the very structure of our learning agent was a significant factor in the robustness of the generated curricula, allowing it to track target trajectories that are significantly different from the trajectories encountered in the curricula. These findings motivate future investigations into the interplay between agent generalization and curricula. From a technical point of view, we demonstrated the importance of appropriately encoding the structure of the context space $\mathcal{C}$ via the distance function of \currot, particularly when the generalization capability of the agent is limited. An interesting next step is to generalize \currot to work with arbitrary Riemannian manifolds. On the robotic side, we see much potential in applications to other robotic tasks, e.g., locomotion problems. On this particular setup, investigating the control of a non-rigidly attached inverted pendulum would allow us to tackle more complicated movements that, e.g., require the robot to thrust the pendulum into the air and catch it again. Furthermore, a non-rigidly attached pendulum would pose an additional challenge for modeling the system in simulation and deriving controllers using optimal control, as contact friction becomes essential to balancing the non-rigidly attached pendulum.

\section*{Acknowledgements}

\noindent Joni Pajarinen was supported by Research Council of Finland (formerly Academy of Finland) (decision 345521).




\appendices

\section{Modelling Network Communication Delays}
\label{app:spherical:network}

\noindent Since our measurements indicated a non-negligible chance of delayed network packets, we modeled this effect during training. With the simulation advancing in discrete timesteps, we modeled the network delays in multiples of simulation steps. More formally, the observation of the pendulum $\svec{x}_{\text{p}, t}$ at time $t$ only becomes available to the agent at time $t + \delta_t$, where $\delta_t \in [0, 1, 2, 3, 4]$. Furthermore, even if $t + i + \delta_{t + i} < t + \delta_{t}$ for some $i > 0$,  the observation at $t + i$ cannot become available before time $t + \delta_{t}$. We realized this behavior by a FIFO queue, where we sample $\delta_t$ upon entry of an observation $\svec{x}_{\text{p}, t}$. \\
We also observed packet losses over the network. Since those losses seemed to correlate with packet delays, we, in each timestep, drop the first packet in the queue with a chance of $25\%$. Hence, the longer the queue is non-empty, i.e., packets are subject to delays, the higher the chance of packets being lost. The probabilities for the delays are given by 
\begin{align}
    p(\delta_t) = \begin{bmatrix} 0.905 & 0.035 & 0.02 & 0.02 & 0.02\end{bmatrix}_{\delta_t}.
\end{align}

\section{Analytic Solution to the LTI System Equations}

\noindent Given that Constraints (\ref{eq:spherical:pos-constraint}) and (\ref{eq:spherical:jerk-constraint}) on the LTI system specify a convex set, which can be relatively easily dealt with, we turn towards  Constraint (\ref{eq:spherical:closing-constraint}), for which we need to derive the closed-form solution of the LTI system (\ref{eq:spherical:lti-sys})
\begin{align}
    \svec{x}(t) = \cvec{\Phi}(t_s, t) \svec{x}(t_s) + \int_{t_s}^t \cvec{\Phi}(\tau, t) \svec{B} u(\tau) \text{d}\tau. \label{eq:spherical:lti-solution}
\end{align}
The transition matrix $\cvec{\Phi}(t_{\change{s}}, t)$ is given by 
\begin{align}
    \cvec{\Phi}(t_s, t) &= e^{\svec{A}\Delta_s} = \svec{I} + \svec{A}\Delta_s + \frac{\svec{A}^2 \Delta_s^2}{2} + \ldots +  \frac{\svec{A}^k \Delta_s^k}{k!} \nonumber \\
    &= \svec{I} + \svec{A}\Delta_s + \frac{\Delta_s^2}{2} \begin{bmatrix} 0 & 0 & 1 \\ 0 & 0 & 0 \\ 0 & 0 & 0 \end{bmatrix} = \begin{bmatrix} 1 & \Delta_s & \frac{\Delta_s^2}{2} \\ 0 & 1 & \Delta_s \\ 0 & 0 & 1 \end{bmatrix}, \label{eq:spherical:lti-phi}
\end{align}
where $\Delta_s = t - t_s$. We can now turn towards the second term in Equation (\ref{eq:spherical:lti-solution}). For solving the corresponding integral, we exploit the assumption that the control signal $u(t)$ is piece-wise constant on the intervals $[t_i,t_{i+1}\change{)}$ with $t_s {=} t_0 {<} t_1 {<} \ldots {<} t_{n-1} {<} t_K {=} t_e$. With that, the second term reduces to
\begin{align}
    \int_{t_0}^t \cvec{\Phi}(\tau, t) \svec{B} u(\tau) \text{d}\tau =\sum_{\change{k}=1}^{K} u_{\change{k}} \int_{t_{\change{k-1}}}^{\min(t_{\change{k}},t)} \cvec{\Phi}(\tau, t) \svec{B}  \text{d}\tau.
\end{align}
We are hence left to solve 
\begin{align}
    \int_{t_l}^{t_h} \cvec{\Phi}(\tau, t) \svec{B} \text{d}\tau &= \int_{t_l}^{t_h} \begin{bmatrix} \frac{(t - \tau)^2}{2} \\ t - \tau \\ 1 \end{bmatrix} \text{d}\tau = \begin{bmatrix} \frac{t^2 \tau}{2} {-} \frac{t \tau^2}{2} {+} \frac{\tau^3}{6} \\ t \tau - \frac{\tau^2}{2} \\ \tau \end{bmatrix} \Biggr|_{\tau=t_l}^{t_h} \nonumber \\
    &= \begin{bmatrix} \frac{t^2 t_h}{2} {-} \frac{t t_h^2}{2} {+} \frac{t_h^3}{6} \\ t t_h - \frac{t_h^2}{2} \\ t_h \end{bmatrix} - \begin{bmatrix} \frac{t^2 t_l}{2} {-} \frac{t t_l^2}{2} {+} \frac{t_l^3}{6} \\ t t_l - \frac{t_l^2}{2} \\ t_l \end{bmatrix} \nonumber \\
    &= \begin{bmatrix} \frac{t^2 \Delta_{lh}}{2} {-} \frac{t \tilde{\Delta}_{lh}^2}{2} {+} \frac{\tilde{\Delta}_{lh}^3}{6} \\ t \Delta_{lh} - \frac{\tilde{\Delta}_{lh}^2}{2} \\ \Delta_{lh} \end{bmatrix} = \change{\cvec{\psi}}(t_l, t_h, t),
\end{align}
where $\Delta_{lh} = t_h - t_l$ and $\tilde{\Delta}_{lh}^i = (t_h - t_l)^i$. \change{We assume $t_l \leq t_h \leq t$ and otherwise define $\cvec{\psi}(t_l, t_h, t)$ to be zero.}

\section{High-Dimensional Ablations}
\label{app:spherical:high_dim}

\begin{figure*}[t]
    \centering
    \includegraphics{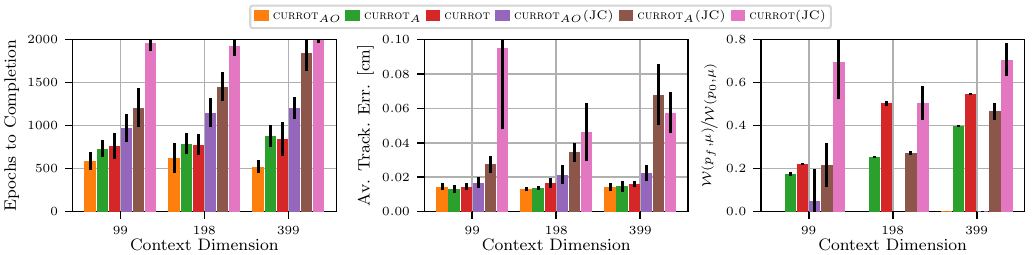}
    \caption{Quantitative results for different \currot versions under increasing dimensions. We show the mean \emph{Epochs To Completion} (left), \emph{Final Average Tracking Error} (middle), and \emph{Final (Normalized) Wasserstein distance} (right). The error bars indicate the standard error. Statistics are computed from $10$ seeds. The abbreviation \textsc{(jc)} stands for experiments in which the agent is given the alternative trajectory representation described in Section \ref{sec:spherical:alternative}.}
    \label{fig:spherical:high_dim_ablation}
\end{figure*}

\noindent To test the robustness of the different \currot versions w.r.t. changes in context space dimensions, we increased the number of sections to represent the jerk trajectory $\svec{u}(t)$. We tested three numbers, resulting in $99$, $198$, and $399$ context space dimensions. When increasing the dimension, we observed that the condition number of the whitening matrix for \newcurrot and \affinecurrot increased significantly, leading to high-jerk interpolations, which were smooth in position and velocity but exhibited strong oscillations in acceleration. \change{We counteracted this behavior by not only measuring the LTI system state via the matrix $\mathbf{A}$ (Section 4.4.1) but also adding the transform $\cvec{\Gamma}_3$ as additional rows to the entries $\cvec{\Psi}_3(t)$ in the definition of $\cvec{\Gamma}$, where $\cvec{\Gamma}$ maps the elements of $\text{ker}(\cvec{\Psi}(t_e))$ to piece-wise constant jerk trajectories and $\cvec{\Gamma}_3$ is its block-diagonal version as defined in the main chapter.} The resulting explicit regularization of the generated jerks prevented the previously observed high jerk interpolations. \\
Figure \ref{fig:spherical:high_dim_ablation} shows the results of the experiments with increasing context space dimensions. The required number of epochs to fully track the target trajectories and the final tracking performance stays almost constant for all methods when using the default trajectory representation, as it allows for good generalization of learned behavior. However, the Wasserstein distances between the final context distribution of the curriculum $p_f(\svec{c})$ and the target distribution $\mu(\svec{c})$ increases with the context space dimension for \affinecurrot and \currot. \\
When using the alternative trajectory representation from Section \ref{sec:spherical:alternative} (indicated by \textsc{(jc)} in Figure \ref{fig:spherical:high_dim_ablation}), this increasingly poor convergence to $\mu(\svec{c})$ leads to a noticeable performance decrease in the required epochs to completion and final tracking performance for \affinecurrot. The performance of \newcurrot decreases only slightly, as the convergence of $p_f(\svec{c})$ to $\mu(\svec{c})$ seems unaffected by higher-dimensional context spaces. As discussed in the main paper, \currot does not allow for good learning with the alternative trajectory representation, rendering the observed tracking performance rather uninformative as they are computed on partially tracked trajectories. The presented results highlight the importance of the improved optimization scheme implemented in \newcurrot, which was not obvious from the experiments in the main paper. 

\bibliographystyle{IEEEtran}
\bibliography{lit.bib}

\end{document}